\documentclass[nonacm, sigconf]{acmart}
\renewcommand\footnotetextcopyrightpermission[1]{} 
\usepackage{hyperref}
\usepackage[hyphenbreaks]{breakurl}
\usepackage{balance}
\usepackage{url}
\usepackage{color}
\usepackage{caption}
\usepackage{diagbox}
\usepackage{subfigure}
\usepackage{multirow}
\usepackage{booktabs}
\usepackage{epsfig,endnotes}
\usepackage{enumitem}
\newcommand{\RNum}[1]{\uppercase\expandafter{\romannumeral #1\relax}}

\newcommand{\para}[1]{{\vspace{2pt} \noindent \textbf{#1}
    \hspace{6pt}}}

\definecolor{applegreen}{rgb}{0.55, 0.71, 0.0}

\newcommand{\eg}{{\em e.g.,\ }}
\newcommand{\ie}{{\em i.e.,\ }}

\newcommand{\secspace}{\vspace{-0.05in}}

\newcommand{\ad}[1]{{$\mathcal{A}$}}
\newcommand{\service}[1]{{$\mathcal{S}$}}

\newcommand{\system}{{\em Gimbal\/}} 

\newenvironment{packed_itemize}{
\begin{list}{\labelitemi}{\leftmargin=0.5em}
  \setlength{\itemsep}{1pt}
  \setlength{\parskip}{0pt}
  \setlength{\parsep}{0pt}
  \setlength{\headsep}{0pt}
  \setlength{\topskip}{0pt}
  \setlength{\topmargin}{0pt}
  \setlength{\topsep}{0pt}
  \setlength{\partopsep}{0pt}
}{\end{list}}

\newenvironment{packed_enumerate}{
\begin{enumerate}
 \setlength{\itemsep}{1pt}
 \setlength{\parskip}{0pt}
 \setlength{\parsep}{0pt}
 \setlength{\headsep}{0pt}
 \setlength{\topskip}{0pt}
 \setlength{\topmargin}{0pt}
 \setlength{\topsep}{0pt}
 \setlength{\partopsep}{0pt}
}{\end{enumerate}}

\begin{document}

\title{Disrupting Style Mimicry Attacks on Video Imagery}
\author{Josephine Passananti$^\dag$, Stanley Wu$^\dag$, Shawn Shan, Haitao Zheng, Ben Y. Zhao\\
$^\dag$ denotes authors with equal contribution\\
  {\em Department of Computer Science, University of Chicago}\\
  {\em \{josephinep, stanleywu, shawnshan, htzheng, ravenben\}@cs.uchicago.edu}}

\begin{abstract}
  Generative AI models are often used to perform mimicry attacks, where a
  pretrained model is fine-tuned on a small sample of images to learn to
  mimic a specific artist of interest. While researchers have introduced
  multiple anti-mimicry protection tools (Mist, Glaze, Anti-Dreambooth),
  recent evidence points to a growing trend of mimicry models using videos as
  sources of training data. 

  This paper presents our experiences exploring techniques to disrupt style
  mimicry on video imagery. We first validate that mimicry attacks can
  succeed by training on individual frames extracted from videos. We show
  that while anti-mimicry tools can offer protection when applied to individual
  frames, this approach is vulnerable to an adaptive countermeasure that removes protection
  by exploiting randomness in optimization results of consecutive
  (nearly-identical) frames. We develop a new, tool-agnostic
  framework that segments videos into short scenes based on frame-level
  similarity, and use a per-scene optimization baseline to remove inter-frame
  randomization while reducing computational cost. We show via both image
  level metrics and an end-to-end user study that the resulting
  protection restores protection against mimicry (including the
  countermeasure). Finally, we develop another adaptive countermeasure and
  find that it falls short against our framework.
\end{abstract}

\maketitle
\pagestyle{plain}

\secspace
\section{Introduction}
\label{sec:intro}

While many debate the ethical and legal issues around the training of
generative image models, all agree that their arrival has dramatically
disrupted a range of visual art industries, from fine art to illustrations,
concept art and graphics arts. Recent works has studied the harms experienced
by professional artists due to these large-scale image generators, including
reputational damage, economic loss, plagiarism and copyright
infringement~\cite{genaiimpact,genaiimpact2}.  One of the more harmful uses
of these image models is ``style mimicry,'' where someone ``finetunes'' a
model on small samples of a specific artist's art, then uses the result to
produce images in the artist's individual style without their
knowledge~\cite{hollie-steal,sarah-andersen,lensa-steal,sam-steal}. These
mimicry models (usually lightweight models known as LORAs) are hosted on
sites including Civitai, Tensor.art, PromptHero and HuggingFace.

Recently, the security and machine learning communities have developed a
number of tools to disrupt unauthorized style mimicry, including
Glaze~\cite{shan2023glaze}, Mist~\cite{mist}, and
Anti-Dreambooth~\cite{antidb}. These tools disrupt the mimicry process, by
modifying images so that they misrepresent themselves in a target model's
style feature space during finetuning, while constraining changes to
minimize visual impact to human eyes.  Since their introduction, they have
been adopted widely by artists across the globe, e.g. Glaze reports more
than 2.3 million downloads in a year~\cite{shan2023glazewebsite}. 

\begin{figure*}[t]
    \centering
    \includegraphics[width=0.75\textwidth]{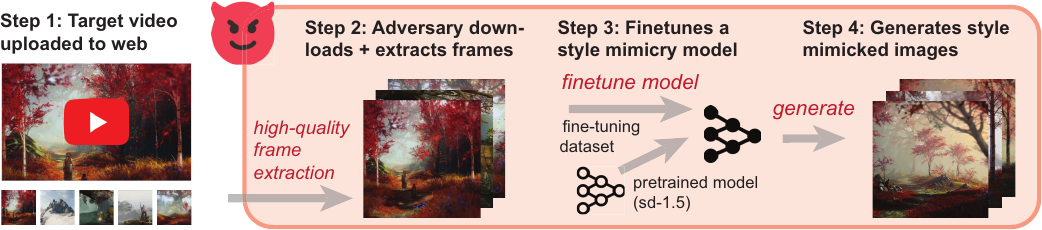}
    \vspace{-0.1in}
    \caption{Style mimicry scenario demonstrating pipeline for adversaries to finetune a diffusion model on video frames}
    \label{fig:style-mimicry-scenario}
  \end{figure*}

There are signs, however, that mimicry attacks are shifting away from 2D art
images and towards video content (see Figure~\ref{fig:style-mimicry-scenario}). Online videos such as animations, game
cut-scenes, music videos and TV shows provide attractive sources
for training mimicry models for several reasons. First, a single video can
provide thousands of frames, each convertible to a standalone image for
training. For example, YouTube videos range from 30 to 60 frames per second,
and even a short 5 minute video can yield 18,000 frames for potential
training. Second, extracting frames from videos provides far more flexibility
to choose a specific scene, character or perspective. This offers far richer
training content compared to static images like movie posters or promotional
art. In fact, many new LORAs already target video games (Riot's League of
Legends\footnote{\url{https://huggingface.co/Totsukawaii/RiotDiffusion}} and
Valorant\footnote{\url{https://huggingface.co/ItsJayQz/Valorant_Diffusion}},
LucasArts
Games\footnote{\url{https://civitai.com/models/270789/lucasarts-games-style}},
Dead Or
Alive\footnote{\url{https://civitai.com/models/382550/kasumi-dead-or-alive-sdxl-lora-pony-diffusion}}),
TV shows (The
Flash\footnote{\url{https://civitai.com/models/42622/danielle-panabaker-the-flash-tv-show}},
Rick \&
Morty\footnote{\url{https://huggingface.co/Madhul/Rick_and_Morty_Stable_Diffusion_LORAS}}),
and movies (Hunger
Games\footnote{\url{https://civitai.com/models/160262/katniss-everdeen-hunger-games}},
Jumanji\footnote{\url{https://civitai.com/models/105883/ruby-roundhouse-from-jumanji-movies-karen-gillan}},
Disney Pixar\footnote{\url{https://tensor.art/models/662818547598142799}}).

The natural question arises: {\em what can we do to protect video creators
  from style mimicry attacks?} This paper presents results from our efforts 
 to answer this question, and to disrupt style mimicry attacks on
video imagery. We begin by first validating the threat. Through empirical
experiments on a variety of short videos, we confirm that it is possible to
produce consistent, high quality mimicry models by extracting and training on
frames from videos. Next, we consider the feasibility of applying existing
tools like Glaze/Mist/Anti-DB to videos on a frame by frame basis. This naive
approach, while computationally expensive, does indeed protect against
extracting and training on video frames.

The problem, however, is that in the video domain, a naive application of
anti-mimicry tools is vulnerable to an adaptive countermeasure. Because
anti-mimicry tools are designed to operate on single images, they compute
protection filters on each single image independently. Randomized components
of these algorithms produce different optimization results on multiple runs
of the same image. For medium to high frame rate videos, this means a clever
attacker can take a protected video, extract consecutive frames whose originals are nearly identical, use the protected frames to identify alterations made by protection tools at the pixel level, and remove them to extract the original frames.  We explore multiple versions of this adaptive attack, and show that they can effectively bypass all 3 anti-mimicry tools and produce mimicry models similar in quality to those trained from
unprotected videos.

Next, we develop \system{}, an anti-mimicry framework for videos to resist this
countermeasure and restore robustness of anti-mimicry tools. At a high level,
we recognize commonalities across all 3 anti-mimicry tools, and extend these
tools to include the notion of sequential similar video frames. Instead of
computing a costly independent protection per frame, we first analyze video
frames to identify scenes or frame sequences with low pixel differential
across frames. The protection tool computes a single baseline perturbation
for all frames in the same scene, and then performs additional local
optimization on a per frame basis. The result is each new frame's protection
pixels is an extension from the prior frame, removing unnecessary
randomization used by the countermeasure. This new approach has the added
benefit of greatly reducing time required to generate protection, often by an
order of magnitude.

We evaluate the efficacy and robustness of \system{} on a variety of
videos. We validate that it integrates naturally with all 3 anti-mimicry
tools. We adapt \system{} to each tool and evaluate the robustness against
video mimicry attacks using multiple image-level metrics, including latent
$L_2$ norm between frames, intra-frame mean 
pixel difference, and CLIP-based genre shift.
Most importantly, we perform a
user study and ask participants to evaluate if the prototype is able to
provide sufficient protection against video mimicry attacks (including the
anti-mimicry countermeasure).  Responses from over 500 participants confirm that
not only does our robust anti-mimicry system works as intended against
mimicry models, but its protection is actually more visually appealing than
naive Glaze (less flickering due to randomization across frames).

Finally, we identify an advanced countermeasure against our frame-aggregation
framework, where an attacker can force us to break a sequence of similar
frames into multiple scenes. We show that even in these scenarios, frame
aggregation prevents an attacker from extracting unprotected frames. 

In summary, our paper makes the following contributions:
\begin{packed_itemize}
  \item We demonstrate that attackers can successfully mimic visual
    styles in videos by extracting and training on individual frames.
  \item We identify and validate the efficacy of an adaptive countermeasure
    that exploits randomization from per-image optimizations to remove image
    protection and enable successful mimicry.
  \item We propose a general anti-mimicry framework for videos that 
    aggregates scenes of similar frames into the protection process, removing
    unnecessary inter-frame randomization, reducing visual artifacts and
    greatly reducing computation costs.
  \item We validate the efficacy of our protection against mimicry attacks
    (including the adaptive countermeasure) using a variety of image metrics
    and two user studies of combined 525 participants.
  \item We propose another adaptive countermeasure on our framework,
    and show that it failures to extract unprotected frames for mimicry.
\end{packed_itemize}

Our work presents an important first step towards protecting video content
against visual style mimicry, by identifying and mitigating a video-specific
countermeasure to anti-mimicry tools. A number of challenges remain, and more
effort is needed to identify other adaptive mimicry algorithms, particularly
for longer videos. 

\section{Background and Related Work}
\label{sec:back}

We begin by providing background on style mimicry and existing image-based protection methods. Then we follow with an overview of publicly available tools that enable style mimicry attacks on the video domain.

\subsection{Style Mimicry and Existing Defenses}
\label{sec:back1}
In a style mimicry attack, a bad actor finetunes a text-to-image model to
generate art in a particular artist's style without their consent. 
Since the introduction of text-to-image diffusion~\cite{sd-release,podell2023sdxl,df,novelai-update,ramesh2022hierarchical} models in 2022, style mimicry has grown significantly. There have been multiple high-profile mimicry incidents involving human artists~\cite{hollie-steal,sarah-andersen,lensa-steal,sam-steal}, and new companies are founded that focus purely on style mimicry~\cite{aigame,lexica}. AI marketplaces have also recently gained traction, with websites like CivitAI~\cite{civitai} offering over 119K ready-to-use mimicry models for people to download and use.

\para{Image-based style mimicry. } 
Style mimicry relies on finetuning pretrained text-to-image models (\eg stable diffusion) on a small set of images from a specific style~\cite{ruiz2022dreambooth,finetune-c,gal2022image}. The quality of these images greatly impacts the mimicry result, and thus, attackers often scrape high quality images from artists' websites and online galleries~\cite{hollie-steal,sam-steal}. In practice, a bad actor does not need many (less than 20 images~\cite{shan2023glaze,gal2022image}) in order to successfully generate arbitrary artwork from a victim artist's style. Because of the risk of image-based mimicry, many artists choose to reduce the amount of art they post online~\cite{aiprotest}, reduce the quality of any posted art~\cite{lowres}, and apply protection (discussed in details below) on this artwork~\cite{shan2023glaze}. 

\para{Protecting images from style mimicry. } 
Existing work (Mist~\cite{mist}, Anti-Dreambooth~\cite{antidb}, and Glaze \cite{shan2023glaze}) has 
proposed methods that leverage clean-label poisoning~\cite{saha2020hidden, turner2018clean, zhu2019transferable} to prevent style mimicry. At a high level, these systems add small optimized perturbations to image artwork that modifies the perturbed image's feature space representation without altering its content. The altered feature space representation prevents models from learning the correct artistic style. In general, these protection tools calculate the perturbation $\delta_x$ for an image $x$ using the following objective: 

\secspace
\begin{eqnarray}
   &\min\limits_{\delta_x} Dist\left( \Phi(x + \delta_x), \Phi(T)\right),  \label{eq:cloakopt}\\
  & \text{subject to } \; |\delta_x|< p, \nonumber
\end{eqnarray} 

where $\Phi$ is a generic image feature extractor from a public text-to-image model, $Dist(.)$ computes the distance between two feature representations, $|\delta_x|$ measures the perceptual perturbation caused by protection, and $p$ is the perceptual perturbation budget. $T$ is 
a ``target image'' that the perturbation $\delta_x$ is optimized towards, such that $x + 
\delta_x$ resembles $T$ in feature space while being visually identical to $x$. Mist~\cite{mist} extends the optimization objective across the entire diffusion process, including gradient computations through the randomized diffusion denoising process. By default, Mist uses a predefined black and white patterned image as its target with the goal of producing chaotic patterns in generated images. Anti-DB (Anti-Dreambooth)~\cite{antidb} is most similar to Mist, but modifies the optimization objective to specifically target Dreambooth~\cite{ruiz2022dreambooth} text-to-image models. There, they find that training surrogate models alongside computing image perturbations results in stronger protection, though it incurs additional computation time. Glaze~\cite{shan2023glaze} introduces input-specific target images by performing style transfer on the input image using a contrasting artistic style. This method preserves the overall content of the input image, while changing mainly the style, which the authors argue leads to more robust protection. Glaze then attacks the image encoder of a diffusion model as detailed above. 

These protection tools have been positively received by the artist community,
with Glaze having been downloaded at least 2.3 million
times~\cite{shan2023glazewebsite}. While these systems are typically too
computationally expensive for artists, efforts have been made to improve
accessibility~\cite{mistgithub, shan2023webglaze}. Since these systems are
free and increasingly available, images may no longer be a viable data source
for attackers to access artwork for fine-tuning text-to-image models. 

Video protection, on the other hand, has yet to be explored. Computation time per image is already limiting for many artists, and applying the same algorithms to all frames would be many times more costly. Yet, videos represent a significant source of data, incentivizing attackers to explore publicly available video art, such as short animation, movies or video game trailers etc..

Until recently, most style mimicry models are trained on still images. This 
is no longer the case today because 1) artists are increasingly more reluctant to post their 
work on the Internet~\cite{aiprotest}, 2) existing defenses (\S\ref{sec:back1}) are effective at protecting still images against mimicry, 3) video frames offer a significantly more diverse range of images compared to still images. 

\para{Video content is a promising source for mimicry. } 
Video content (\eg game trailers, anime, short videos, documentary, ads) provides promising alternative data sources for two reasons. First, video contents often offer a more diverse (3D) shots of an object or style, \eg rotating shot of an object, panning across a scene. These diverse viewpoints 
enable models to better learn the content during the training process~\cite{videomotivate}. Second, there are significantly more video frames compared to still images and many of the videos contain unique art styles/characters. The entire Internet produces around 3.2 billions still images daily~\cite{manyvideos}, while YouTube alone sees over 271,000 hours of videos (\ie around 29 billions video frames) uploaded per day. Specifically, gaming companies and animation studios often use short videos as a way to promote new games, characters, and movies. Movie clip compilations and trailers are readily available on YouTube~\cite{movietrailers}, while video game companies like Riot and Mihoyo frequently post teasers and trailers showcasing new playable content, or highly anticipated characters~\cite{genshintrailer, leaguetrailer}. These videos are filled with original artwork, and contain image frames that are prime targets for style mimicry. 

\para{Video-based mimicry in the real-world. } 
Style mimicry using video content has already occurred in the real-world. Bad actors have created and distributed software that generates high quality text-to-image datasets from online videos. One GitHub tool~\cite{anime2sd} automates the process of downloading (\eg torrenting) Japanese Anime episodes and extracting high quality frames of desired characters. Another option~\cite{civitai-video} advertised on CivitAI does the same, with the additional capability of scraping frames from screencap websites such as FanCaps~\cite{fancaps}. These tools demonstrate that there already exists sophisticated technology aimed at creating text-to-image datasets from original video content. 

\begin{figure}[t]
    \centering
    \includegraphics[width=1\columnwidth]{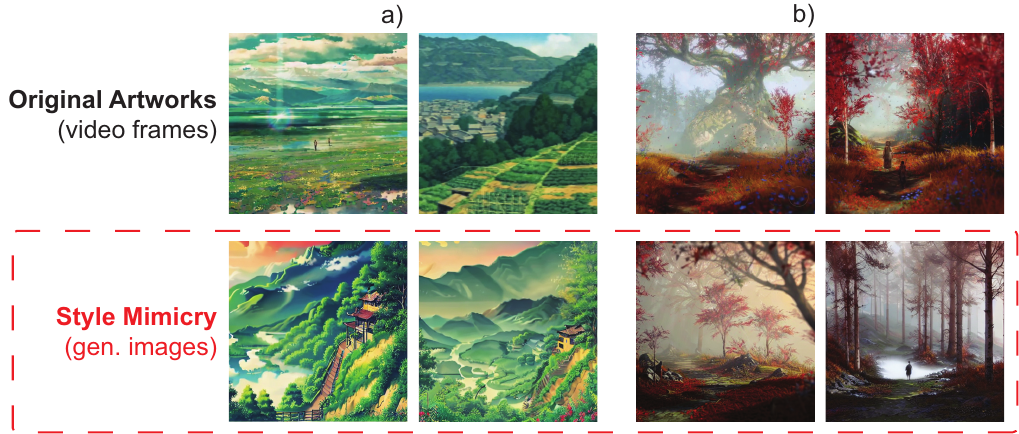}
    \vspace{-0.2in}
    \caption{Style mimicry on clean video frames successfully mimics style of original video.}
    \label{fig:style-mimicry-baseline}
  \end{figure}

We also provide our own examples of this threat. We download and extract high quality frames from YouTube videos and train style mimicry models on them (Figure~\ref{fig:style-mimicry-scenario}). Figure~\ref{fig:style-mimicry-baseline} shows some examples of extracted video frames as well as mimicry results generated by the style mimicry model. We include human evaluation of the success of these style mimicry images later using user studies with both artists and the general public in \S\ref{sec:eval}. 

While there has been recent developments in text-to-video~\cite{videoworldsimulators2024} and image-to-video models~\cite{blattmann2023stable}, we leave them as a topic for future work, and focus solely on text-to-image mimicry where the source of data originates from video content.

\secspace
\section{Style Mimicry Attacks on Extracted Video Frames}
\label{sec:threat}

Our work considers a previously overlooked variant of the style mimicry
attack, where an attacker extracts individual frames from a video to build
image-based style mimicry models. Next, we introduce the threat model and
consider the limitations of a baseline defense that applies existing
image-based protection tools to individual video frames.

\secspace
\subsection{Threat Model}

\para{Artist/Video creator. } Artists want to share their video 
content online while disallowing unauthorized mimicry using these video frames. 
Artists seeks to protect their video by applying small 
pixel perturbations on the video frames. Following the assumptions made by existing defenses against style mimicry~\cite{shan2023glaze}, we
assume the artists: 
\begin{packed_itemize} 
    \item have access to moderate computing resources (\eg consumer-grade GPUs) commonly used for video rendering; 
    \item add perturbation to video frames before posting videos online; 
    \item have access to some public feature extractor (\eg open-source models such as Stable Diffusion).
\end{packed_itemize}

\para{Attacker. } The attacker's goal is to build a \textit{text-to-image} model
that is able to generate images in the style of the victim artist. We assume the 
attacker 
\begin{packed_itemize} 
\item has access to videos from the victim and leverages frames from
these videos for mimicry; 
\item has significant computational power; 
\item has full access to pretrained, benign text-to-image base models. 
\end{packed_itemize}
Note that our work focuses on text-to-image mimicry, where the attacker's goal
is to {\bf generate images}. We leave text-to-video 
mimicry using video contents to future work. 

\subsection{Style Mimicry Leveraging Video Frames}

\subsection{A Naive Defense and Its Limitations}
\label{subsec:limitations}

Given the existence of existing tools designed to disrupt art style
mimicry~\cite{shan2023glaze,mist,antidb}, a straightforward solution to
protect video imagery is to simply apply existing protection to each and
every frame of a video.  As discussed in \S\ref{sec:back}, these defenses add
highly-optimized perturbations on each image (a video frame in our case),
misleading the mimicry model to perceive each protected frame as an image with a
completely different style.

Unfortunately, this ``naive'' application of anti-mimicry tools in the video
context has multiple drawbacks, the most critical of which is vulnerability
to a temporal-based adaptive countermeasure.

\para{Vulnerability to countermeasures based on temporal similarity.}  When
applied on individual video frames without coordination, existing protection
methods become vulnerable to advanced countermeasures that exploit
visual similarity across consecutive video frames.

Existing anti-mimicry tools treat each input image independently, with a
randomized component in their generation of perturbation targets for each
image. That means even two runs on the same input image will likely output
two different set of pixel changes. With this in mind, an adaptive mimicry
attack could take several consecutive frames, whose original pixel values are
highly similar, and use them against each other to try to cancel out the
pixel changes made by the protection tools.  An``averaged'' or
``smoothed'' frame generated this way would have much weaker residue
protective perturbations, and would provide a good estimate of the
actual visual feature (i.e. style) carried by the original (unperturbed)
video frames.  An attacker can then use these frames to train a
mimicry model. In \S\ref{sec:eval-limitations}, we provide a detailed study to
validate and quantify this significant vulnerability.

\para{Other limitations: computation and video quality degradation.} The
naive application of protection tools leads to two other challenges. First,
computing independent protection filters on each video frame is
computationally very expensive.  Existing protection tools
(Glaze~\cite{shan2023glaze}, Mist~\cite{mist} and Anti-DB~\cite{antidb}) can
takes up to $\tau=1.5$ minutes to protect a single frame for moderate
GPUs. Protecting a 1 minute video at 30fps (1800 frames) would take 45 hours.
Second, the protective pixel level perturbations are computed independently per
frame and hard to detect on a still image. But when played in a video
sequence, these perturbations cause noticeable flickering effects that
degrade the video quality and viewer experience.

\secspace
\section{An Adaptive Mimicry Attack} 
\label{sec:eval-limitations}
In this section, we design, implement and evaluate an adaptive mimicry attack
designed to bypass the naive protection method described in
\S\ref{subsec:limitations}. 

It is made possible by the fundamental {\em temporal consistency} inherent to
all video content. Here, we propose {\em Perturbation Removal Attacks} (PRA)
that leverage temporal consistency to remove protective perturbations on
video frames, and present results measuring their efficacy using both
automated metrics and human feedback.

\secspace
\subsection{Perturbation Removal}

For naively protected video sequences, we develop an adaptive mimicry attack
that uses perturbation removal methods (PRM) to recover images that closely
approximate the original, unperturbed video frames. These are then used to
successfully train an image-based style mimicry model.  As such, for any
video sequence, a PRMs behaves like a frame extraction tool.

\para{``Combining'' consecutive frames.} PRMs remove protective perturbations
by combining multiple consecutive video frames (that share high visual
similarity) into a single frame. Intuitively, combining highly similar frames
will generally preserve the common pixel values inherited from the original
unprotected frames, while reducing or removing the pixel value changes made
by protective tools.  With this in mind, we consider three PRM approaches that
employ different ``combining'' functions across video frames.

\begin{packed_itemize}
\item \textbf{Selective Pixel Averaging} -- This approach generates a
  combined image out of consecutive frames, where each pixel is the average
  of the corresponding pixel values across the set of frames. This
  pixel-level ``averaging'' function ``smooths'' out the protective
  perturbations. Pixel averaging can be limited to more static regions and
  avoid pixels that capture motion across the frames. 
\item \textbf{FILM} -- Frame Interpolation for Large Motion
  (FILM)~\cite{reda2022film} is a neural network designed to generate
  temporally smooth videos from disjoint frames. We apply FILM to multiple
  perturbed frames from the same scene to reconstruct high quality images
  that closely resemble the original unperturbed frames, but don't retain
  perturbations from either input.
\item \textbf{Linear Interpolation} -- Linear interpolation is
  a technique for generating intermediate data points between a set of known
  points. Like the other approaches, pixel-level linear interpolation 
  across frames exploits lack of consistency between consecutive
  perturbations.
\end{packed_itemize}

\para{Implementing adaptive attacks.} We implemented 3 adaptive mimicry attacks, each
using one of the frame-aggregation approaches described above. In the rest of
this section, we present detailed results on all 3 adaptive attacks. The key
takeaway is that pixel averaging significantly outperforms the
alternatives. For brevity, we move implementation details of the other two
attacks to Appendix \ref{app:detailed-perturbation-removal}, and only provide
implementation details for the pixel averaging below. 

\begin{figure}[t]
  \centering
  \includegraphics[width=1\columnwidth]{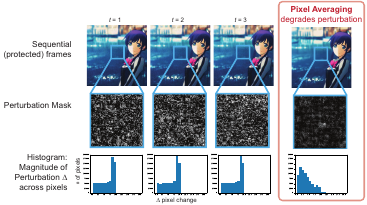}
  \vspace{-0.2in}
  \caption{Averaging pixel values across highly similar consecutive frames
    successfully degrades the randomized protection pixel shifts across
    frames and largely restores the original unprotected frame. }
  \label{fig:pixel-averaging-attack}
\end{figure}

\emph{Pixel Averaging} approximates the original unprotected frame by
averaging pixels across highly similar consecutive frames (see
Figure~\ref{fig:pixel-averaging-attack}). Pixel level differences between
consecutive frames come from two sources: 1) natural changes between video
frames which we call \textit{movement}, and 2) differences in pixel changes
added by protective tools (\textit{perturbations}). Protective tools seek to
minimize visual impact, so the large majority of perturbation values are
constrained within a specific value. Thus an attacker can examine two
consecutive perturbed frames, and identify the source of each pixel
difference by filtering using a simple threshold ($\epsilon_p$). A well
chosen $\epsilon_p$ will separate pixel differences due to movement
($>\epsilon_p$) from pixel differences from protective tools
($<\epsilon_p$). The attacker measures pixel differences ($\Delta_p$) between
consecutive frames and only averages regions ($0 < \Delta_p <
\epsilon_p$). In practice, $\epsilon_p$ can easily be identified empirically
as a transition point between two levels of region sizes. In our tests, we
experimentally test pixel averaging across $n$ consecutive frames, and find
the best results around $n=5$.  We measure the quality of frames using CLIP
Aesthetic predictor~\cite{schuhmann2022laion} and perturbation removal using
metrics described in the following section. We show detailed results of the
tradeoff between perturbation removal vs. image quality in the appendix.

\subsection{Experimental Setup and Metrics}
To validate the efficacy of multiple perturbation removal methods, we add
naive protection to short scenes on an independent, per-frame basis. We then
test the adaptive mimicry attack by using each PRM to extract unprotected
frames from each video scenes. We compare different PRMs by measure the
amount of image level differences in the perturbations before vs. after our
attack using several automated metrics. We also measure end-to-end
success of the adaptive mimicry attack by training models on
extracted frames, and conducting a user study to gather human
feedback.

\para{Applying naive protection to all frames.}  We experiment on 5 diverse
datasets containing realistic videos of scenery and human actions, artistic
style videos, and video game style videos. We implement each of 3 protection
tools, Mist, Anti-DB, and Glaze.  For each scene, we identify and extract
scenes of highly similar frames, and apply ``naive protection'' by applying
each of 3 protection tools to each frame in selected frames. We apply each
PRM to the naively protected images to attempt to recover a good estimation
of the original images. We then compare original, naively protected, and
attacked naively protected images to each other.
As we show below, pixel-averaging significantly outperforms FILM and linear
interpolation in pixel level metrics.

\para{Performing style mimicry.} We perform style mimicry attacks under 3
``perturbation scenarios'': training mimicry models on 
clean (unperturbed) frames, frames protected by ``naive protection,'' and
frames extracted by adaptive attack following ``naive protection.''  Due to
high computation costs (multiple days per scene), we only compute end-end
results for the Pixel Averaging attack (shown to be strongest in pixel level
metrics above).
For each perturbation scenario, we chose ~30
scenes from a video, select (or extract) one image from each scene,
and train mimicry models on this set.  Further details on mimicry attack
configurations are in \S\ref{sec:eval}.

\para{Evaluation metrics.}  We evaluate the strength of perturbation removal
methods using pixel level metrics (Mean Pixel Difference and Latent $L_2$ Norm). 
We evaluate end-to-end results on style mimicry using human feedback
(User Study). We briefly describe these metrics below, and give more details later in
\S\ref{sec:eval}.

\begin{packed_itemize}
\item{\em Latent $L_2$ Norm.} We employ the image encoder used in diffusion models
  to calculate image representations of perturbed and non-perturbed (original)
  frames, and then calculate the $L_2$ distance between them as a measure
  of proximity between two images. Thus, a successful perturbation removal
  would minimize the latent $L_2$ norm, while a robust system should maintain a
  high latent $L_2$ norm.

\item{\em Mean Pixel-Difference.}  We measure differences between images at a
  pixel level, motivated by the $l_{inf}$ bounded pixel changes that all
  protection tools (Mist, Anti-DB, Glaze) use to limit visual artifacts. We
  define Mean Pixel Difference (MPD) as the average of all pixel differences
  between a perturbed image and clean image. Similar to the latent $L_2$ norm, 
  a higher MPD signals higher protection.

\item{\em Human feedback.}  We perform a user study to evaluate the success
  of adaptive mimicry attacks, by asking participants to look at images
  produced by a mimicry model, and compare it to original video frames.  We
  ask participants to rate the success on a 5-level Likert scale (ranging
  from ``not successful at all'' to ``very successful''). Following existing
  work, we define protection success rate (PSR) as the percent of
  participants who rated the style mimicry as ``not very successful'' or
  ``Not successful at all.''
\end{packed_itemize}

\begin{figure}[t]
  \centering
  \includegraphics[width=1\columnwidth]{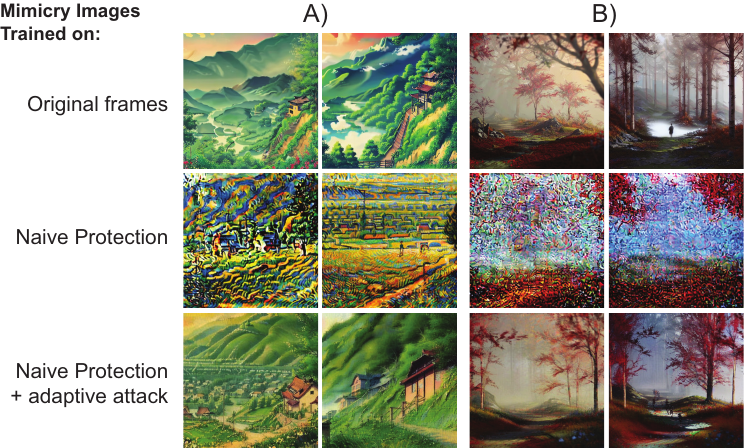}
  \vspace{-0.3in}
  \caption{Visual examples of adaptive mimicry attack. Three rows of
      mimicry images generated by models trained on 1) original video frames,
    2) video frames protected naively, and 3) video frames recovered after
    perturbation removal using pixel averaging.}
  \label{fig:style-mimicry-attacked}
\end{figure}

\begin{table}[t]
  \centering
    \resizebox{0.5\textwidth}{!}{
    \centering
\begin{tabular}{c|cccc}
 Protection Tool  & Protected          & Pixel Avg                   & FILM Interpolation & Linear Interpolation \\ \hline
  Glaze          & 390.05 $\pm$ 25.17 & \textbf{295.55 $\pm$ 52.02} & 391.51 $\pm$ 62.86 & 355.58 $\pm$ 44.38   \\
  Mist           & 474.87 $\pm$ 42.14 & \textbf{334.62 $\pm$ 56.72} & 437.78 $\pm$ 64.89 & 411.27 $\pm$ 51.89   \\
  Anti-DB        & 405.17 $\pm$ 37.55 & \textbf{283.41 $\pm$ 58.99} & 378.43 $\pm$ 67.34 & 352.10 $\pm$ 52.30  
  \end{tabular}
  }\caption{Latent $L_2$ Norm between original frames, protected frames
    and protected frames after perturbation removal.}
\label{tab:loss-removal-results}
\vspace{-0.2in}
\end{table}

\begin{table}[t]
  \centering
    \resizebox{0.5\textwidth}{!}{
    \centering
\begin{tabular}{c|cccc}
  Protection Tool & Protected          & Pixel Avg                   & FILM Interpolation & Linear Interpolation \\ \hline
  Glaze          & 111.30 $\pm$ 12.94 & \textbf{90.04 $\pm$ 14.13}  & 97.83 $\pm$ 13.73  & 96.16 $\pm$ 14.47    \\
  Mist           & 121.25 $\pm$ 9.48  & \textbf{107.10 $\pm$ 16.33} & 112.99 $\pm$ 14.36 & 113.03 $\pm$ 15.46   \\
  Anti-DB        & 124.86 $\pm$ 8.93  & \textbf{103.75 $\pm$ 17.70} & 110.29 $\pm$ 15.52 & 110.39 $\pm$ 16.42  
  \end{tabular}
  }\caption{Mean Pixel Difference between original frames, protected frames,
    protected frames after perturbation removal.}
\label{tab:pd-removal-results}
\vspace{-0.2in}
\end{table}

\subsection{Adaptive Mimicry Results}

Next we present results of our experiments on the adaptive attack.

\para{Similarity of recovered frames to original frames.} We compare
{\em latent $L_2$ norm} 
between the original frames, the protected frames, and protected frames after
protection removal.  Table~\ref{tab:loss-removal-results} shows FILM to
have minimal impact, and that Pixel averaging does the best to minimize loss
for the recovered frame, suggesting that it is closer to the original frame
in the feature space.

We also compare {\em mean pixel differences} between the original
frames, protected frames, and protected frames after protection removal.
Table~\ref{tab:pd-removal-results} shows that again, pixel averaging method
outperforms against all protection tools, and minimizes the pixel differences
between the extracted frame and the original. 

\para{Style mimicry attack on recovered images.}  Finally, we use a user
study to evaluate the end-to-end success of the adaptive mimicry attack using
pixel-averaging to overcome a per-frame application of protection tools.
Table~\ref{tab:user-study-removal-results} shows that users agree, the
adaptive mimicry attack with pixel averaging basically produces mimicry
results similar to mimicry on original unprotected video frames (PSR baseline
value of 17.65 for unprotected video frames).
Figure~\ref{fig:style-mimicry-attacked} shows samples of mimicry images from
models trained on original frames, protected frames, and protected frames
under adaptive attack. Clearly the adaptive attack is able to bypass
protection and restore mimicry success.

These results validate our concerns, that a naive, frame by frame application
of protection tools to videos is insufficient to prevent image mimicry. We
need to extend these anti-mimicry tools to restore their protection in the
video domain.

\secspace

\section{Protecting Video Imagery with \system{}}
\label{sec:method}
We have identified that video imagery is susceptible to style mimicry attacks
despite existing protection in image space (\S\ref{sec:eval-limitations}).
Although current protection tools are effective for 2D art, they are no longer robust
to adaptive adversaries in the video domain. In this section, we
develop \system, a framework that extends image-based protection tools to the video
domain, resulting in improved robustness against adaptive adversaries as well
as lower computation costs and improved video quality for protected videos.

\begin{table}[t]
  \centering
    \resizebox{0.4\textwidth}{!}{
    \centering
\begin{tabular}{c|cc}
  Protection Tool & Protected        & Protected + Pixel Averaging                 \\ \hline
  Glaze          & 70.59 $\pm$ 1.05 & \textbf{23.76 $\pm$ 1.14} \\
  Mist           & 62.90 $\pm$ 1.11 & \textbf{25.34 $\pm$ 1.14} \\
  Anti-DB        & 59.28 $\pm$ 1.18 & \textbf{22.62 $\pm$ 1.15} 
  \end{tabular}}
  \caption{Human feedback (Protection Success Rate) shows perturbation
    removal can significantly reduce effects of protection tools 
    against style mimicry attacks. Note baseline PSR for original,
    unprotected frames is 17.65 $\pm$ 1.10.}
\label{tab:user-study-removal-results}
\vspace{-0.2in}
\end{table}

\begin{figure*}[t]
  \centering
  \includegraphics[width=1\textwidth]{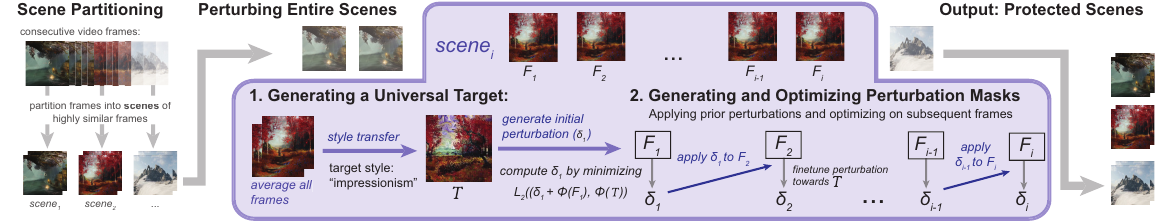}
  \vspace{-0.25in}
  \caption{\system~ partitions videos into scenes by measuring average pixel difference. Each scene is perturbed using a two-part process: 1) A target image ($T$) is computed by averaging all frames and style transferring to a 'target style' 2) Perturbations are iteratively applied and optimized by minimizing latent $L_2$ norm between $T$ and the perturbed frame.}
  \label{fig:system-design}
\end{figure*}

\subsection{Design intuition} 

\para{Challenges of existing perturbation systems.} Existing perturbation
systems do not take into account the threat of an adversary gaining access to
highly similar or even identical frames that are protected with completely
independent perturbations.  These systems independently optimize separate
protection perturbations for each frame.

This optimization process includes 1) generating a \textit{random} target latent
tensor, and 2) perturbing the original image by optimizing the 
image towards the selected target.  The choice of target significantly
impacts the pixel level perturbations on an image. Current systems choose
targets by generating a latent representation of the original image,
introducing noise, and then employing denoising autoencoders.  This
randomness (\ie non-linearity) is desired for image-based protection, making
it more challenging to reverse engineer or remove the perturbations. However,
it also leads to each perturbation mask being entirely unique; disregarding
the temporal redundancy present in the original frames. This in turn gives
rise to effective countermeasures that remove the protection
(\S\ref{sec:eval-limitations}).

\para{Intuition.}
Our key intuition is that if we can create very similar perturbations on
similar underlying frames, we can nullify the adversaries ability to exploit
duplicity in frames. With this in mind, there are two simple options for
perturbing a scene of similar frames. The first option is to re-use the same
perturbation for the entire scene. Re-using the same perturbation is robust
to pixel averaging attacks, but frame-specific protection weakens after
small levels of movement in frames.

A second option is that if we can chose very similar targets (that guide the
optimizing of perturbations) for similar frames, we can increase the
consistency between perturbations. We find that re-using a single target
tensor throughout a scene leads to very similar perturbations, but also makes
frames less robust to style mimicry. This is because the perturbation
optimization algorithm is not able to customize the embeddings of different
frames to the same degree as it does for the original frame the
target tensor was generated from. Thus our goal is to 1) divide 
videos into scenes that can share a \textit{universal} target, 2) generate
this ``universal target'' for each scene, and 3) optimize perturbations on
subsequent frame to maximize protection against mimicry.

\subsection{System Design}

Figure~\ref{fig:system-design} describes our pipeline for generating robust,
protected videos by partitioning them into scenes, generating a target image
for each scene, and then optimizing the perturbations on each frame towards
its respective target. 

\para{Scene partitioning.}  We want to pinpoint sections of videos where
using different perturbations might leave frames vulnerable to averaging
attacks. We split videos into distinct scenes based on frame similarity, and
considered existing scene partitioning algorithms and
tools~\cite{pyscenedetect}. Note that we require all frames in a scene to be
similar enough to share a single ``perturbation target.'' This is a stronger
constraint than most prior definitions for a ``scene,'' leading us to
implement our own algorithm. We split scenes based on the mean pixel
difference between two consecutive frames $F_i$ and $F_{i-1}$, defined
by \secspace
\begin{equation}
 \frac{Pixel(F_i, F_{i-1})}{N}< \epsilon_{scene}\label{eq:mean_pixel_diff}
\end{equation}
where $N$ is the number of pixels per frame and $Pixel(.)$ calculates
the pixel difference between two frames, and $\epsilon_{scene}$ is a
parameter for scene partition. 

\para{Generating a universal target.}  To create consistency between
consecutive perturbations, we compute a single target image for all frames in
a scene. This means that the target embedding needs to be close enough the
embeddings of each frame in the scene, so that it can correctly guide each
individual optimization. We test several approaches to generating the target
tensor: 1) using the middle frame from the scene as base image, 2) generating
image embeddings of each frame in the scene, calculating the centroid of
these embeddings as base image, and 3) averaging all frames together as base
image. We measure latent $L_2$ norm distance between the embedding of each unique frame in a
scene and the target tensor, and find that averaging all frames in a scene
leads to a target that is consistently the closest distance across all frames
(Figure~\ref{fig:target-selection-algorithm-results} in Appendix).

Specifically, we compute a style transferred target image $T$ from the averaged
image, like prior work~\cite{shan2023glaze}. We select a video-specific
prompt to guide the style transfer, for example: \textit{Japanese Anime} videos
use ``impressionist painting by Van Gogh'' as the target style.

\para{Generating and optimizing perturbation masks.}  We consider several
factors when computing perturbations: maximizing protection, robustness
against perturbation removal attacks, and finally, reducing computation
costs. We balance re-using perturbations on subsequent frames \textit{which
  enhances robustness against removal attacks} with optimizing or recomputing
perturbations \textit{which maintains high protection levels against mimicry
  attacks}. Protection success is achieved by reducing loss between a
perturbed image and target image. We develop our algorithm for perturbing a
scene as follows:

\begin{packed_itemize}
  \item Given the current scene and its corresponding $M$ frames $\{F_i\}_{i=1}^{M}$, compute the target frame $T$ (described above).
  \item For the first frame $F_1$, compute an image-based perturbation
    $\delta_{i}$ from scratch, optimizing towards the target frame
    $T$, as defined by eq. (\ref{eq:cloakopt}). 
   \item For each subsequent frame $i$ ($i=2..M$), first compute
     \begin{equation}
       d_i=|L_2(\Phi(F_i+\delta_{i-1}), \Phi(T)) -
       L_2(\Phi(F_{i-1}+\delta_{i-1}),\Phi(T))| \label{eq:vg}
     \end{equation}
     where $\Phi(.)$ is 
     the feature extractor used to convert an image into a
     latent embedding, $L_2$ is the L2
     distance between the two embeddings. Compute  $\delta_{i}$, the perturbation for
     $F_i$  as follows:
     
     \begin{itemize}
     \item $d_i\leq\tau_1$: reuse the previous frame's perturbation,
       i.e. $\delta_{i}=\delta_{i-1}$;
     \item $\tau_1<d_i\leq\tau_2$: compute $\delta_{i}$ by
      performing perturbation optimization towards $T$ starting from
      $\delta_{i-1}$;
      \item $d_i > \tau_2$, compute $\delta_{i}$ from scratch.
    \end{itemize}

\end{packed_itemize}
Here we note that because the
  perturbation is generated progressively using the same target image
  $T$, the system is able to maintain 
  high consistency between perturbations.

\para{Setting $\tau_1$ and $\tau_2$.}
\label{subsec:system-parameters}
Our system applies two thresholds $\tau_1$ and
$\tau_2$ to guide the perturbation computation.  We use grid search
to identify proper values that balance robustness and computational
efficiency in all videos. In practice, content creators can perform a benchmark on their own videos to select thresholds that best balance their robustness and efficiency requirements. Further details on our grid search is located in the Appendix.

\secspace
\section{Evaluation} 
\label{sec:eval}
In this section, we evaluate \system's ability to protect individual video
frames from style mimicry. \S\ref{sec:setup} describes our video
datasets and experimental setup. \S\ref{sec:eval-metrics} introduces
our metrics for evaluation. \S\ref{sec:protection-robustness}-\ref{sec:video-types}
present results on \system's protection and efficiency. Due to the subjective
nature of interpreting successful style mimicry and visual nature of videos,
we evaluate protection using both automated metrics from existing work, as
well as visual judgement in a user study.

\para{Summary of results.}
Naive perturbations protect videos against style mimicry attempts 64.3\% of
the time according to surveyed users. However, perturbation removal attacks
on naive perturbations successfully recover video style, causing protection
rate to drop to 23.8\% (compared to completely unprotected videos at
17.7\%). \system{} is able to maintain protection even against removal
attacks, restoring protection against style mimicry attempts to 64.5\%. 
We verify that \system~ is able to preserve image perturbations across
consecutive image frames on a diverse set of videos genres and types, and is
robust against our averaging attack introduced in \S\ref{sec:eval-limitations}. 

\subsection{Experimental Setup} 
\label{sec:setup}
\para{Video datasets.}
We evaluate the effectiveness of \system~ on five diverse datasets, covering
different animations, as well as human actions and scenery. On average, each
video contains 6289 total frames. Our experiments conducted on YouTube videos
are for research purposes only, and trained models are deleted at the
conclusion of the study~\cite{fairuse}.
\begin{enumerate}
\item \textbf{Video Games}: 20 randomly selected YouTube videos from a single
  channel showcasing in-game content of different video games, ranging from 2
  to 6 minutes long.
\item \textbf{Human Actions}: 20 randomly selected untrimmed videos of human
  actions from the THUMOS-15 training dataset originally designed for action
  classification~\cite{Idrees_2017}, ranging from 1 to 4 minutes long.
\item \textbf{Japanese Anime}: 20 randomly selected YouTube videos from
  different channels of Japanese animated movies and tv-shows, ranging from 1
  to 5 minutes long.
\item \textbf{Animated Movies}: 10 randomly selected compilations of animated
  (Disney, Pixar, Sony) movie clips, ranging from 5 to 13 minutes long.
\item \textbf{Nature and Wildlife}: 10 randomly selected YouTube videos of
  animals and scenery in the wild, shot in a documentary format. Original
  videos are 30-60 minutes long, we clip each video to the first 2 minutes.
\end{enumerate}

We include three unique animation styles, Video Game (3D
rigs~\cite{videogameanimation}), Japanese Anime (hand
drawn~\cite{animeanimation}), and Animated Movies
(CGI~\cite{amoviesanimation}). The remaining two video datasets are selected
to cover the other types of video imagery, including real human presence
and photography/aerial footage. As a preprocessing step, we maintain
consistent quality among all videos by center cropping videos to square and
resizing to 512x512.  

\para{Defense configuration.}  To showcase the generalizability of \system~
to multiple image-based protection systems, we test \system~ on three such
algorithms: Mist, Anti-DB, and Glaze. We re-implement each defense using
$l_{\infty}$ bounded projected gradient descent (PGD), and use a consistent
targeted image generation method for all three~\cite{shan2023glaze}. As is
standard in image-based PGD methods, we constrain maximum absolute change
in each image pixel to 0.07.

\para{Mimicry configuration.}  We base our style mimicry setup on existing
work~\cite{mist,shan2023glaze,shan2023prompt,antidb} and adapt it to video
frames:
\begin{packed_enumerate}
\item Split each video into partitions using our scene splitting algorithm from \S\ref{sec:method}.
\item Use the CLIP aesthetic model to identify the frame with the highest image quality score within a scene, and select the highest quality images based on scenes with the highest image quality score.
\end{packed_enumerate}
    
We find that all datasets could successfully train style mimicry models with
30 images, except for Animated Movies, which required 60. Here, we can apply the pixel 
averaging adaptive mimicry attack, finetuning on 80\% of extracted images
and saving the remaining 20\% for testing. We use Dreambooth to
finetune a Stable Diffusion 2.1 model on the finetuning dataset for 1000
steps and learning rate of 1e-5. To generate matching synthetic images, we
generate captions from testing images with BLIP~\cite{li2022blip}, and query
the finetuned model to create a set of style mimicked images. 

We evaluate \system~ across several combinations of our proposed defense
system and style mimicry configurations. For each existing perturbation
algorithm, we evaluate the efficacy when perturbed images are untouched, and
when our image averaging attack is used. We evaluate the efficacy of \system~
against style mimicry only when the image averaging attack is used. We also
train a style mimicry model on every video where the frames are left
untouched.

\subsection{Evaluation Metrics}
\label{sec:eval-metrics}
The two key contributions of \system~ are robustness under potential
adversary attack, and reduction in computation costs, which we will evaluate
using the following metrics.

\para{Robustness.}  We measure robustness in two ways. The first two metrics
capture how well image perturbations can withstand adversarial attack. The
last two measure the impact that \system~ has on the quality of images
generated by style mimicry models.
\begin{packed_itemize}
\item \textbf{Latent $L_2$ Norm:} We employ the image encoder used in
  diffusion models to calculate latent representations of perturbed and
  non-perturbed (original) frames, and then calculate the $L_2$ distance
  between them as a measurement for image closeness. For example, a
  successful averaging attack on consecutively perturbed frames would lead to
  low latent $L_2$ norm, while a robust system should maintain a high latent
  $L_2$ norm. In particular, this metric isolates how diffusion models
  capture image differences. 
\item \textbf{Mean Pixel Difference:} We also measure the differences between
  images at a pixel level, motivated by the $l_{\infty}$ bounded pixel
  changes that all three of our defense algorithms are constrained by. Mean
  Pixel Difference (MPD) is model-agnostic, and defined as the average of all
  pixel differences between a perturbed image and original image. Similar to
  the latent $L_2$ norm, a higher MPD between perturbed and original frames
  signals higher protection.
\item \textbf{CLIP-Genre Shift:} We adapt a metric used in existing
  work~\cite{shan2023glaze} to demonstrate the effectiveness
  of \system~ at disrupting style mimicry. Intuitively, existing image
  perturbation algorithms are designed to cause diffusion models to learn the
  wrong artistic style. Thus, we can measure the success of \system's
  protection by calculating the percentage of generated images in which a
  CLIP image classifier unsuccessfully predicts the ground-truth style from
  training images. A higher CLIP-based Genre Shift score equates to stronger
  protection, while the opposite equates to stronger style mimicry. However, it 
  is a granular metric and its correlation to visual properties might be weak.
\item \textbf{Human-rated protection success rate (PSR):} To accurately capture 
  end-to-end visual properties, we perform
  two IRB-approved user studies (more details on participants in
  Appendix~\ref{app:detailed-user-study}) where participants look at images
  generated by 
  mimicry models and rate if the style mimicry was successful. The first asks
  artist volunteers to 
  rate performance on our two ``Art'' datasets: Anime and Video Games. The
  second asks general users to rate performance on all 5 datasets. For
  each video, we train 10 style mimicry models, matching our experiment
  configuration. For each mimicry model, we show participants 5 frames from
  the original video next to 5 frames generated by the mimicry model. Each
  participant is shown one example from each experiment configuration,
  randomly selected from the set of 80 videos (3 datasets x 20 videos + 2
  datasets x 10 videos each).

  \indent We ask participants to compare original
  video frames to those generated by mimicry, and rate the success on a 5-point
  Likert scale (ranging from ``Not successful at all'' to ``Very
  successful''). Following prior work, we define protection success rate as
  the percent of participants who rated how well generated images mimic
  original style as ``Not very well'' or ``Not well at all.'' We also show
  artists short 10 second clips of videos glazed naively and with
  \system. Here, we ask how noticeable the perturbations are on a 5 point
  Likert scale ranging from ``Very noticeable'' to ``Not noticeable at all.''
  We define \textit{Noticeability Rate} (NR) as the \% of users that think
  perturbations are ``Noticeable'' or ``Very noticeable.''
\end{packed_itemize}

\para{Computation efficiency:} 
Finally, we also measure the computation speed of \system. Here, we only experiment on one image perturbation algorithm due to computational constraints. Glaze is chosen due to its balance of speed (fastest of the three) and strong robustness. For each video, we conduct our measurement on a single A100 GPU.
\begin{packed_itemize}
\item \textbf{Speedup Factor:} We compute speedup factor as time taken to apply (naive)
  protection to every frame, divided by time taken to protect the video using
  \system.  Because of the compute time involved, we estimate full (naive) protection by estimating per frame protection time, scaled up by number of frames in the video.
\item \textbf{Seconds per Frame:} We also report the average time it takes to
  perturb each frame in a video. This provides a grounding metric that is not
  relative like speedup factor.
\end{packed_itemize}

\begin{table*}[t]
    \centering
    \resizebox{0.88\textwidth}{!}{
      \begin{tabular}{cccccccccc}
        & \multicolumn{3}{c}{Glaze}                                                                                                                                               & \multicolumn{3}{c}{Mist}                                                                                                                                                & \multicolumn{3}{c}{Anti-DB}                                                                                                                        \\ \cline{2-10} 
        & Naive              & \begin{tabular}[c]{@{}c@{}}Naive\\ + Attack\end{tabular} & \multicolumn{1}{c|}{\textbf{\begin{tabular}[c]{@{}c@{}}\system\\ + Attack\end{tabular}}} & Naive              & \begin{tabular}[c]{@{}c@{}}Naive\\ + Attack\end{tabular} & \multicolumn{1}{c|}{\textbf{\begin{tabular}[c]{@{}c@{}}\system\\ + Attack\end{tabular}}} & Naive              & \begin{tabular}[c]{@{}c@{}}Naive\\ + Attack\end{tabular} & \textbf{\begin{tabular}[c]{@{}c@{}}\system\\ + Attack\end{tabular}} \\ \hline
\multicolumn{1}{c|}{Video Game}          & 406.69 $\pm$ 20.09 & 262.82 $\pm$ 32.39                                       & \multicolumn{1}{c|}{425.19 $\pm$ 26.84}                                                 & 468.51 $\pm$ 40.68 & 296.29 $\pm$ 38.53                                       & \multicolumn{1}{c|}{496.04 $\pm$ 43.52}                                                 & 400.48 $\pm$ 35.12 & 240.38 $\pm$ 37.83                                       & 421.66 $\pm$ 37.01                                                 \\
\multicolumn{1}{c|}{Japanese Anime}      & 395.86 $\pm$ 20.32 & 284.51 $\pm$ 53.59                                       & \multicolumn{1}{c|}{406.93 $\pm$ 26.87}                                                 & 467.00 $\pm$ 43.44 & 319.77 $\pm$ 56.92                                       & \multicolumn{1}{c|}{493.53 $\pm$ 47.91}                                                 & 406.04 $\pm$ 41.29 & 272.34 $\pm$ 59.09                                       & 424.51 $\pm$ 41.96                                                 \\
\multicolumn{1}{c|}{Animated Movies}     & 380.04 $\pm$ 17.52 & 308.06 $\pm$ 47.66                                       & \multicolumn{1}{c|}{406.30 $\pm$ 26.05}                                                 & 475.07 $\pm$ 39.40 & 349.04 $\pm$ 51.88                                       & \multicolumn{1}{c|}{496.81 $\pm$ 42.17}                                                 & 404.55 $\pm$ 37.85 & 299.84 $\pm$ 55.27                                       & 422.11 $\pm$ 38.46                                                 \\
\multicolumn{1}{c|}{Nature and Wildlife} & 402.10 $\pm$ 17.24 & 319.23 $\pm$ 53.86                                       & \multicolumn{1}{c|}{408.62 $\pm$ 25.91}                                                 & 466.05 $\pm$ 41.64 & 346.28 $\pm$ 56.05                                       & \multicolumn{1}{c|}{496.48 $\pm$ 41.38}                                                 & 393.73 $\pm$ 30.02 & 297.58 $\pm$ 58.51                                       & 412.73 $\pm$ 30.82                                                 \\
\multicolumn{1}{c|}{Human Actions}       & 370.77 $\pm$ 27.41 & 316.53 $\pm$ 50.39                                       & \multicolumn{1}{c|}{397.97 $\pm$ 27.03}                                                 & 494.19 $\pm$ 39.40 & 369.70 $\pm$ 48.57                                       & \multicolumn{1}{c|}{508.08 $\pm$ 39.06}                                                 & 415.92 $\pm$ 36.58 & 316.19 $\pm$ 50.97                                       & 426.26 $\pm$ 35.29                                                
\end{tabular}
    }
    \caption{Latent $L_2$ norm between original and perturbed frames across all datasets and image perturbation algorithms. Averaging attack reduces perturbation effectiveness from naive video cloaking (attacked naive $<<$ naive), but is unsuccessful against \system~.}
    \label{tab:adv-algorithm-robustness-loss}
\end{table*}

\begin{table*}[t]
    \centering
    \resizebox{0.88\textwidth}{!}{
      \begin{tabular}{cccccccccc}
        & \multicolumn{3}{c}{Glaze}                                                                                                                                               & \multicolumn{3}{c}{Mist}                                                                                                                                                & \multicolumn{3}{c}{Anti-DB}                                                                                                                        \\ \cline{2-10} 
        & Naive              & \begin{tabular}[c]{@{}c@{}}Naive\\ + Attack\end{tabular} & \multicolumn{1}{c|}{\textbf{\begin{tabular}[c]{@{}c@{}}\system\\ + Attack\end{tabular}}} & Naive              & \begin{tabular}[c]{@{}c@{}}Naive\\ + Attack\end{tabular} & \multicolumn{1}{c|}{\textbf{\begin{tabular}[c]{@{}c@{}}\system\\ + Attack\end{tabular}}} & Naive              & \begin{tabular}[c]{@{}c@{}}Naive\\ + Attack\end{tabular} & \textbf{\begin{tabular}[c]{@{}c@{}}\system\\ + Attack\end{tabular}} \\ \hline
\multicolumn{1}{c|}{Video Game}          & 111.99 $\pm$ 11.50 & 85.00 $\pm$ 12.37                                        & \multicolumn{1}{c|}{108.54 $\pm$ 12.66}                                                 & 120.88 $\pm$ 6.79  & 102.91 $\pm$ 14.20                                       & \multicolumn{1}{c|}{119.26 $\pm$ 7.52}                                                  & 124.64 $\pm$ 5.94  & 98.21 $\pm$ 14.69                                        & 121.10 $\pm$ 7.49                                                  \\
\multicolumn{1}{c|}{Japanese Anime}      & 112.31 $\pm$ 12.92 & 91.03 $\pm$ 14.44                                        & \multicolumn{1}{c|}{108.25 $\pm$ 14.26}                                                 & 120.84 $\pm$ 8.71  & 106.25 $\pm$ 15.08                                       & \multicolumn{1}{c|}{119.03 $\pm$ 9.88}                                                  & 124.34 $\pm$ 7.94  & 103.01 $\pm$ 15.91                                       & 120.27 $\pm$ 10.27                                                 \\
\multicolumn{1}{c|}{Animated Movies}     & 109.43 $\pm$ 13.14 & 89.92 $\pm$ 12.16                                        & \multicolumn{1}{c|}{107.57 $\pm$ 13.56}                                                 & 122.22 $\pm$ 8.59  & 109.44 $\pm$ 15.08                                       & \multicolumn{1}{c|}{121.37 $\pm$ 9.19}                                                  & 125.89 $\pm$ 7.98  & 106.27 $\pm$ 15.99                                       & 122.42 $\pm$ 9.21                                                  \\
\multicolumn{1}{c|}{Nature and Wildlife} & 115.14 $\pm$ 5.39  & 98.65 $\pm$ 12.56                                        & \multicolumn{1}{c|}{109.96 $\pm$ 8.23}                                                  & 121.54 $\pm$ 3.13  & 110.49 $\pm$ 10.65                                       & \multicolumn{1}{c|}{120.80 $\pm$ 4.81}                                                  & 124.79 $\pm$ 2.30  & 108.15 $\pm$ 13.53                                       & 120.92 $\pm$ 5.25                                                  \\
\multicolumn{1}{c|}{Human Actions}       & 109.61 $\pm$ 16.02 & 90.18 $\pm$ 15.91                                        & \multicolumn{1}{c|}{108.22 $\pm$ 16.41}                                                 & 120.85 $\pm$ 14.49 & 108.19 $\pm$ 21.66                                       & \multicolumn{1}{c|}{119.06 $\pm$ 16.02}                                                 & 124.52 $\pm$ 14.13 & 105.51 $\pm$ 23.66                                       & 120.09 $\pm$ 16.64                                                
\end{tabular}
    }
    \caption{Mean pixel difference (MPD) between original and perturbed frames across all datasets and image perturbation algorithms. Averaging attack reduces perturbation effectiveness from naive video cloaking (attacked naive $<<$ naive), but is unsuccessful against \system~.}
    \label{tab:adv-algorithm-robustness-pd}
\end{table*}

\begin{table*}[t]
    \centering
    \resizebox{0.88\textwidth}{!}{
      \begin{tabular}{cccccccccccc}
        \multirow{2}{*}{}                                                        & \multirow{2}{*}{}                         &                                       & \multicolumn{3}{c}{Glaze}                                                                                                                                             & \multicolumn{3}{c}{Mist}                                                                                                                                              & \multicolumn{3}{c}{Anti-DB}                                                                                                                      \\ \cline{3-12} 
                                                                                 &                                           & \multicolumn{1}{c|}{Clean}            & Naive            & \begin{tabular}[c]{@{}c@{}}Naive\\ + Attack\end{tabular} & \multicolumn{1}{c|}{\textbf{\begin{tabular}[c]{@{}c@{}}\system\\ + Attack\end{tabular}}} & Naive            & \begin{tabular}[c]{@{}c@{}}Naive\\ + Attack\end{tabular} & \multicolumn{1}{c|}{\textbf{\begin{tabular}[c]{@{}c@{}}\system\\ + Attack\end{tabular}}} & Naive            & \begin{tabular}[c]{@{}c@{}}Naive\\ + Attack\end{tabular} & \textbf{\begin{tabular}[c]{@{}c@{}}\system\\ + Attack\end{tabular}} \\ \hline
        \multirow{2}{*}{Artist}                                                  & \multicolumn{1}{c|}{Video Game}           & \multicolumn{1}{c|}{15.05 $\pm$ 1.10} & 92.23 $\pm$ 0.72 & 19.35 $\pm$ 1.13                                         & \multicolumn{1}{c|}{97.98 $\pm$ 0.52}                                                   & 82.88 $\pm$ 0.85 & 14.81 $\pm$ 1.08                                         & \multicolumn{1}{c|}{91.01 $\pm$ 0.69}                                                   & 93.26 $\pm$ 0.69 & 23.71 $\pm$ 1.25                                         & 91.74 $\pm$ 0.67                                                   \\
                                                                                 & \multicolumn{1}{c|}{Japanese Anime}       & \multicolumn{1}{c|}{24.27 $\pm$ 1.13} & 82.88 $\pm$ 0.97 & 27.03 $\pm$ 1.13                                         & \multicolumn{1}{c|}{72.32 $\pm$ 1.00}                                                   & 68.69 $\pm$ 0.97 & 34.74 $\pm$ 1.18                                         & \multicolumn{1}{c|}{65.77 $\pm$ 1.10}                                                   & 79.44 $\pm$ 0.83 & 36.19 $\pm$ 1.14                                         & 83.84 $\pm$ 0.83                                                   \\ \hline
        \multirow{5}{*}{\begin{tabular}[c]{@{}c@{}}General\\ Users\end{tabular}} & \multicolumn{1}{c|}{Video Game}           & \multicolumn{1}{c|}{20.14 $\pm$ 1.13} & 85.59 $\pm$ 0.82 & 18.26 $\pm$ 1.11                                         & \multicolumn{1}{c|}{88.99 $\pm$ 0.76}                                                   & 86.84 $\pm$ 0.87 & 16.67 $\pm$ 1.06                                         & \multicolumn{1}{c|}{95.24 $\pm$ 0.65}                                                   & 73.87 $\pm$ 1.02 & 25.20 $\pm$ 1.17                                         & 80.39 $\pm$ 0.97                                                   \\
                                                                                 & \multicolumn{1}{c|}{Japanese Anime}       & \multicolumn{1}{c|}{20.95 $\pm$ 1.07} & 66.07 $\pm$ 1.05 & 21.21 $\pm$ 1.09                                         & \multicolumn{1}{c|}{54.63 $\pm$ 1.07}                                                   & 59.46 $\pm$ 1.18 & 26.50 $\pm$ 1.14                                         & \multicolumn{1}{c|}{68.80 $\pm$ 1.03}                                                   & 49.00 $\pm$ 1.07 & 19.61 $\pm$ 0.99                                         & 52.53 $\pm$ 1.21                                                   \\
                                                                                 & \multicolumn{1}{c|}{Animated Movies}      & \multicolumn{1}{c|}{14.81 $\pm$ 1.19} & 44.83 $\pm$ 1.15 & 14.81 $\pm$ 1.04                                         & \multicolumn{1}{c|}{36.51 $\pm$ 1.16}                                                   & 35.00 $\pm$ 1.04 & 16.67 $\pm$ 1.06                                         & \multicolumn{1}{c|}{50.91 $\pm$ 1.19}                                                   & 26.79 $\pm$ 1.14 & 12.24 $\pm$ 1.03                                         & 35.94 $\pm$ 1.12                                                   \\
                                                                                 & \multicolumn{1}{c|}{Nature and Wildlife}  & \multicolumn{1}{c|}{8.89 $\pm$ 1.02}  & 78.43 $\pm$ 0.80 & 18.06 $\pm$ 1.10                                         & \multicolumn{1}{c|}{65.38 $\pm$ 1.02}                                                   & 68.63 $\pm$ 0.98 & 20.41 $\pm$ 1.07                                         & \multicolumn{1}{c|}{85.00 $\pm$ 0.73}                                                   & 69.49 $\pm$ 1.16 & 31.34 $\pm$ 1.34                                         & 64.15 $\pm$ 1.01                                                   \\
                                                                                 & \multicolumn{1}{c|}{Generic Video Scenes} & \multicolumn{1}{c|}{15.05 $\pm$ 1.03} & 70.64 $\pm$ 1.05 & 42.42 $\pm$ 1.18                                         & \multicolumn{1}{c|}{49.02 $\pm$ 1.17}                                                   & 54.00 $\pm$ 1.05 & 38.74 $\pm$ 1.15                                         & \multicolumn{1}{c|}{60.00 $\pm$ 1.06}                                                   & 62.73 $\pm$ 1.15 & 21.51 $\pm$ 1.19                                         & 61.21 $\pm$ 1.20                                                  
        \end{tabular}
    }
    \caption{Percentage of \textit{artists} and \textit{general users} who
      deem protection is successful on different videos. Artists rated
      more art-driven videos (Video Games and Japanese Anime), while general
      users rated all videos. Column include naive protection,
      pixel-averaging on naive protection, and pixel-averaging on
      \system{}. Pixel-averaging breaks protection; \system{} restores it.}
    \label{tab:adv-algorithm-robustness-artist}
\end{table*}

\begin{table*}[t]
    \centering
    \resizebox{0.88\textwidth}{!}{
      \begin{tabular}{ccccccccccc}
        &                                      & \multicolumn{3}{c}{Glaze}                                                                                                                                            & \multicolumn{3}{c}{Mist}                                                                                                                                             & \multicolumn{3}{c}{Anti-DB}                                                                                                                     \\ \cline{2-11} 
        & \multicolumn{1}{c|}{Clean}           & Naive           & \begin{tabular}[c]{@{}c@{}}Naive\\ + Attack\end{tabular} & \multicolumn{1}{c|}{\textbf{\begin{tabular}[c]{@{}c@{}}\system\\ + Attack\end{tabular}}} & Naive           & \begin{tabular}[c]{@{}c@{}}Naive\\ + Attack\end{tabular} & \multicolumn{1}{c|}{\textbf{\begin{tabular}[c]{@{}c@{}}\system\\ + Attack\end{tabular}}} & Naive           & \begin{tabular}[c]{@{}c@{}}Naive\\ + Attack\end{tabular} & \textbf{\begin{tabular}[c]{@{}c@{}}\system\\ + Attack\end{tabular}} \\ \hline
\multicolumn{1}{c|}{Video Games}         & \multicolumn{1}{c|}{0.49 $\pm$ 0.23} & 0.85 $\pm$ 0.23 & 0.71 $\pm$ 0.26                                          & \multicolumn{1}{c|}{0.97 $\pm$ 0.07}                                                    & 0.85 $\pm$ 0.20 & 0.75 $\pm$ 0.22                                          & \multicolumn{1}{c|}{0.97 $\pm$ 0.06}                                                    & 0.77 $\pm$ 0.26 & 0.61 $\pm$ 0.26                                          & 0.94 $\pm$ 0.13                                                    \\
\multicolumn{1}{c|}{Japanese Anime}      & \multicolumn{1}{c|}{0.25 $\pm$ 0.18} & 0.76 $\pm$ 0.23 & 0.58 $\pm$ 0.19                                          & \multicolumn{1}{c|}{0.88 $\pm$ 0.13}                                                    & 0.77 $\pm$ 0.24 & 0.60 $\pm$ 0.21                                          & \multicolumn{1}{c|}{0.92 $\pm$ 0.11}                                                    & 0.64 $\pm$ 0.30 & 0.41 $\pm$ 0.23                                          & 0.86 $\pm$ 0.13                                                    \\
\multicolumn{1}{c|}{Animated Movies}     & \multicolumn{1}{c|}{0.45 $\pm$ 0.22} & 0.65 $\pm$ 0.29 & 0.52 $\pm$ 0.28                                          & \multicolumn{1}{c|}{0.68 $\pm$ 0.25}                                                    & 0.61 $\pm$ 0.28 & 0.53 $\pm$ 0.30                                          & \multicolumn{1}{c|}{0.73 $\pm$ 0.21}                                                    & 0.58 $\pm$ 0.31 & 0.46 $\pm$ 0.29                                          & 0.65 $\pm$ 0.28                                                    \\
\multicolumn{1}{c|}{Nature and Wildlife} & \multicolumn{1}{c|}{0.32 $\pm$ 0.15} & 0.80 $\pm$ 0.24 & 0.61 $\pm$ 0.20                                          & \multicolumn{1}{c|}{0.91 $\pm$ 0.12}                                                    & 0.82 $\pm$ 0.18 & 0.53 $\pm$ 0.30                                          & \multicolumn{1}{c|}{0.98 $\pm$ 0.02}                                                    & 0.69 $\pm$ 0.32 & 0.41 $\pm$ 0.22                                          & 0.94 $\pm$ 0.09                                                    \\
\multicolumn{1}{c|}{Human Actions}       & \multicolumn{1}{c|}{0.22 $\pm$ 0.23} & 0.70 $\pm$ 0.32 & 0.50 $\pm$ 0.31                                          & \multicolumn{1}{c|}{0.78 $\pm$ 0.22}                                                    & 0.80 $\pm$ 0.25 & 0.66 $\pm$ 0.27                                          & \multicolumn{1}{c|}{0.94 $\pm$ 0.07}                                                    & 0.68 $\pm$ 0.33 & 0.45 $\pm$ 0.27                                          & 0.81 $\pm$ 0.24                                                   
\end{tabular}
  }
  \caption{CLIP-Genre Shift across all datasets and image perturbation algorithms. Averaging attack decreases the number of images generated of a different style to the original video (attacked naive $\approx$ clean), but fails to do the same against \system~.}
    \label{table:adv-algorithm-style-mimicry}
\end{table*}

\subsection{Robustness against Pixel-Averaging Mimicry}
\label{sec:protection-robustness}
\para{\system~ prevents style mimicry. } 
We showed in \S\ref{sec:eval-limitations} that pixel-averaging attacks can
remove image protection by smoothing them out across similar
frames. We begin by looking at the ability of pixel-averaging methods to extract
frames similar to the original, as measured by pixel level metrics in
Table~\ref{tab:adv-algorithm-robustness-loss} and
\ref{tab:adv-algorithm-robustness-pd}.

For all protection tools, across each category of videos, we see that the
pixel-averaging attack significantly reduces the distance between the
protected frames compared to the originals, measured by both latent $L_2$ norm and
MPD. More importantly, we see that that same pixel-averaging attack fail when
applied to frames protected by \system{}, and it actually increases the
distances from the original.

Next, we turn our attention to the ability of adaptive mimicry attacks and
their ability to produce accurate end-to-end mimicry models. Table~\ref{tab:adv-algorithm-robustness-artist} shows the results of our
two user studies, involving both a population of artists and general users
(participant details in Appendix~\ref{app:detailed-user-study}). While
artists' views varied somewhat from general users across categories, all
users consistently provided the same feedback, that pixel-averaging broke the
protection provided by naive anti-mimicry tools, but \system{} restored that
protection (and in many cases increased protection higher than naive
protection levels). Table~\ref{table:adv-algorithm-style-mimicry} quantified
the same issue of end to end protection, but using the CLIP
CLIP-Genre shift metric. Results are very consistent with those from user
studies. \system{} restored protection broken by pixel-averaging attack and
in many cases, improved protection beyond the original naive levels.

\begin{figure}[t]
    \centering
    \includegraphics[width=0.95\columnwidth]{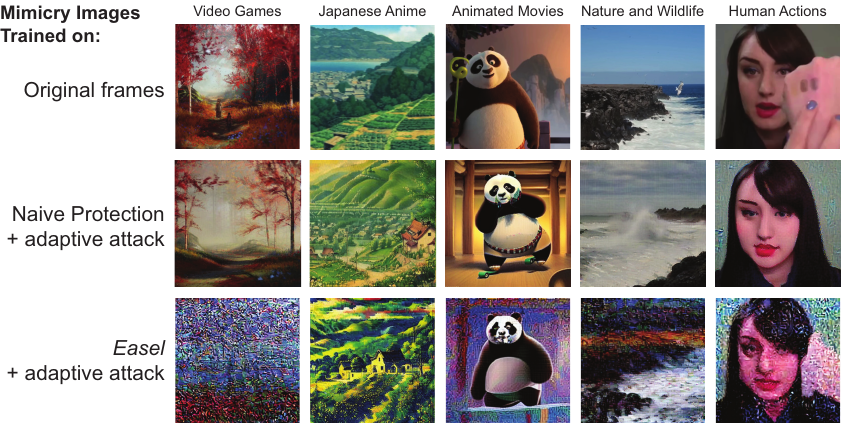}
    \vspace{-0.1in}
    \caption{Some visual examples of style mimicry on \system~ showing it is
      robust to pixel averaging attack.}
    \label{fig:core-style-mimicry-results}
    \vspace{-0.1in}    
  \end{figure}

Finally, Figure~\ref{fig:core-style-mimicry-results} shows some examples of
images generated by style mimicry models under our experiment configurations.

\begin{figure}[t]
    \centering
    \includegraphics[width=3in]{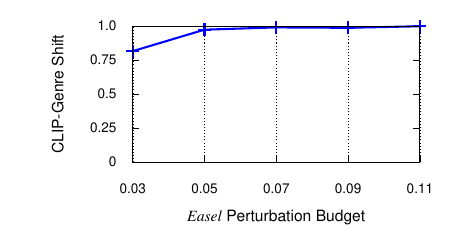}
    \vspace{-0.2in}
    \caption{Average robustness (CLIP-Genre Shift) increases as $l_{\infty}$
      perturbation budget increases on 4 Video Game videos. For comparison,
      mimicry on original frames produces CLIP-Genre Shift of 0.44.}
    \label{fig:eps-robustness}
\end{figure}

\para{Impact of perturbation budget on robustness.} 
In Figure~\ref{fig:eps-robustness}, we show that robustness under \system{}
increases with perturbation budget, quickly maxing out after 0.05. This
follows the same trend we see in existing image-only perturbation algorithms.

\subsection{Computational Costs and Video Quality}
\label{subsec:protection-usability}
In \S\ref{subsec:limitations}, we identified two additional limitations
of a naive application of anti-mimicry tools: high computation overhead and
poor video quality (randomness across per-frame perturbations appear as
flickering snow when viewed at regular speeds). Here, we show that
\system~ significantly improves on both the computation overhead of
perturbing video frames, and increases video quality over naively protected
videos.

\para{\system~ reduces computation overhead. } 
We measure the computation efficiency of \system~ on three of our
datasets. As shown in Table~\ref{tab:adv-algorithm-efficiency}, the Video
Game and Human Actions datasets achieve 7-8x speedup factor when using \system~ over
its naive video protection, while the Japanese Anime dataset achieves just over
4x speedup factor. \system~ would improve computation efficiency of a 5 minute 30
fps video from 88 to 11 hours. This magnitude of speedup factor greatly improves the
usability of applying image-based perturbations to videos, and makes video
protection more reasonable for video creators, particularly smaller or
independent creator groups.

The speedup factor is lowest for Japanese anime videos, likely due to the high movement 
across frames. In practice, we can further improve the speedup by changing our
system parameters, \eg changing second threshold $\tau_2$
from 0.45 to 0.8, which increases speedup factor of the four slowest Japanese Anime videos from 2.12 to 3.90 with a small tradeoff in robustness (more details
in Appendix~\ref{app:num-images-average} and \ref{app:eps-robustness}).  

\begin{table}[t]
    \centering
    \resizebox{0.4\textwidth}{!}{
        \begin{tabular}{c|cc}
            & Avg. Speedup Factor    & Avg. Seconds per Frame \\ \hline
            Video Game     & 7.87 $\pm$ 2.81 & 5.16 $\pm$ 2.13        \\
            Japanese Anime & 4.16 $\pm$ 3.19 & 11.25 $\pm$ 4.55       \\
            Human Action   & 7.37 $\pm$ 4.34 & 7.84 $\pm$ 6.43       
        \end{tabular}
    }
    \caption{\system~ with Glaze enables significant computation speedup factor
      across all Video Game, Japanese Anime, and Human action videos when
      compared to naive video protection. For comparison, naive video
      protection takes on average 35 seconds per frame on a single A100 GPU.}
    \label{tab:adv-algorithm-efficiency}
\end{table}

\begin{table}[t]
    \centering
    \resizebox{0.4\textwidth}{!}{
        \begin{tabular}{c|cc}
            & \multicolumn{2}{c}{\% of Users Notice Perturbations} \\
            & Naive Glaze               & \system~ w/ Glaze           \\ \hline
            Video Game     & 70.30 $\pm$ 1.03          & 21.29 $\pm$ 1.06         \\
            Japanese Anime & 48.73 $\pm$ 1.16          & 18.27 $\pm$ 1.0         
        \end{tabular}
    }
    \caption{Our user study shows that the percentage of users who notice
      perturbations generated by \system~ is
      significantly less than those who notice perturbations in naive video
      protection. (Tests implemented using Glaze).} 
    \label{tab:adv-algorithm-style-mimicry-visual}
\end{table}

\para{\system~ is less visually disruptive/noticeable. }  Even though
image-based perturbation methods are designed to be imperceptible, applying
them naively to videos leads to protected videos that flicker and are
noticeably grainy. This is a result of perturbation masks changing
drastically from frame to frame, a problem that \system~ addresses
directly. In our user study, we show participants 10-second clips of
protected videos from our three datasets. The first video is naively protected
using our implementation of Glaze, and the second video is protected with 
\system{}. Participants are asked to identify how visible the
perturbations are in both 
videos. Table~\ref{tab:adv-algorithm-style-mimicry-visual} presents
\textit{noticeability rate}, showing artists are able to notice perturbations
significantly more when videos are protected naively (70\%) than with
\system~ (21\%).

\begin{table}[t]
    \centering
    \resizebox{0.3\textwidth}{!}{
        \begin{tabular}{c|cc}
            & CLIP-Genre Shift & Speedup Factor          \\ \hline
            15fps & 0.97 $\pm$ 0.04  & 6.11 $\pm$ 1.90  \\
            30fps & 0.98 $\pm$ 0.02  & 9.57 $\pm$ 2.47  \\
            60fps & 1.00 $\pm$ 0.00  & 15.96 $\pm$ 4.06
        \end{tabular}
    }
    \caption{Different framerates do not impact protection
      robustness (CLIP-Genre Shift), but higher framerates lead to higher
      speedup factor under \system. (4 Video Game videos.)} 
    \label{table:fps-test}
\end{table}

\begin{figure}[t]
  \centering
  \includegraphics[width=0.9\columnwidth]{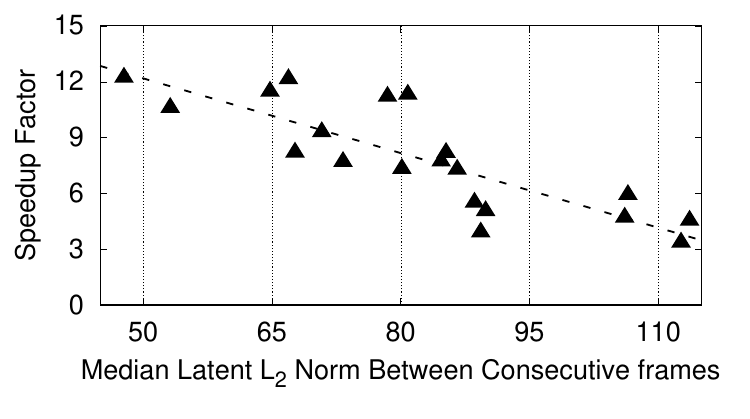}
  \vspace{-0.25in}
  \caption{Speedup factor decreases as movement increases between frames (latent $L_2$
    norm) on all Video Game videos.}
  \label{fig:action-across-datasets}
\end{figure}

\subsection{Impact of Video Types on Robustness and Efficiency}
\label{sec:video-types}
Beyond examining the efficacy of \system~ on a variety of video genres, we
also examine four aspects of videos creation/distribution irrespective to
genre that real-world content creators are likely to deal with while posting
videos online. We investigate the effects that framerate, movement between
frames, duration of scenes, and compression, have on robustness and
computation efficiency. We anticipate these as questions that video content
creators are likely to ask with respect to the efficacy of \system. In this
section, we show that \system's robustness holds steady across video types,
while computational efficiency is the most prone to change.

\para{FPS}
Here, we evaluate the change in robustness and computation efficiency of
\system~ when videos are encoded with a different number of frames per second
(fps). We evaluate \system~ with Glaze on three different framerates, 15fps,
30fps, and 60fps, and measure its performance on four videos from the Video
Game Dataset. 30fps and 60fps are widely used in the video
community~\cite{ytvideosettings}, and we also include 15 to show the effects
that much a slower framerate has on robustness and computation
efficiency. Our results can be found in Table~\ref{table:fps-test}, which
show that framerate does not have significant impact on robustness, but that
increasing framerate from 15fps to 60fps increases speedup factor from 6x to
16x. Thus, we show that \system's computation performance is at its worst
with low framerate, but scales well when content creators choose to increase
framerate in their videos.

\para{Action within scenes}
Movement within videos is difficult to capture, but we use latent $L_2$ norm between
two image latent representations as a good approximation. Intuitively, minor
movement between frames should lead to minor changes in image latent
representation. In this section, we investigate the effect that latent $L_2$
norm between consecutive frames in video scenes has on computation
efficiency. We measure the latent $L_2$ norm between consecutive frames in a
scene for every video in the Video Game dataset, and compare the median of
the latent $L_2$ norm across scenes to the speedup factor \system~ with Glaze
provides. In Figure~\ref{fig:action-across-datasets}, we show that there is a
negative linear correlation between $L_2$ latent norm and
speedup factor. Intuitively, videos with higher action within scenes are more likely
to require additional optimization in order to better align image
perturbations across fast changing frames. 

Finally, we also analyze the impact of {\bf scene duration} (number of frames per
scene) and {\bf video compression} factor (bitrate of video) on robustness and
speedup factor. We found that these factors have no observable impact.

\section{Scene Splitting Adaptive Attack}
\label{sec:counter}

\system~ removes uncontrolled randomness in perturbations between similar
frames, thereby disabling adaptive attacks that leverage cross frame pixel
correlations. However, does its own design give rise to new countermeasures?
We carefully considered this question, and discuss what we consider the
strongest possible countermeasure to \system{}. We describe the potential
countermeasure and evaluate its efficacy against \system.

The intuition for the countermeasure is to manipulate the scene
identification process to force a scene break between highly similar
frames. If this can be achieved, then the attacker could obtain two
consecutive (and similar) frames that have been perturbed towards different
targets. They could then apply a version of the pixel-averaging attack to
restore the original, unprotected frames. We call this the scene-splitting
attack, and assume a powerful attacker who can somehow force \system{} to
insert a scene break in the middle of a sequence of similar frames.

\para{The scene splitting attack.} 
We simulate a strong adaptive attack by dividing a single scene from a
video into two subscenes, Scene $S_1$ and $S_2$ each containing $M$ frames. We
use each subscene (Scene$_i$) to generate a target tensor $T_{i}$ with the
aim of maximizing distance between target tensors, which should maximize the
difference between perturbations generated from these targets. 
Now that we have obtained $T_1, T_2$, we apply \system~ to perturb the two
consecutive frames around the bad scene split. Subsequently, we perform a
pixel averaging attack on these two perturbed frames, following the
implementation detailed in \S\ref{sec:eval-limitations}. 

We test our adaptive attack on 20 videos from the Japanese Anime and Video
Game datasets (10 videos from each dataset). We choose these two datasets
because they are best aligned with current threats as explained in
\S\ref{sec:intro}. We limit ourselves to only 10 videos from each category
because of the computation time involved.

\begin{figure}[t]
  \centering
  \includegraphics[width=1\columnwidth]{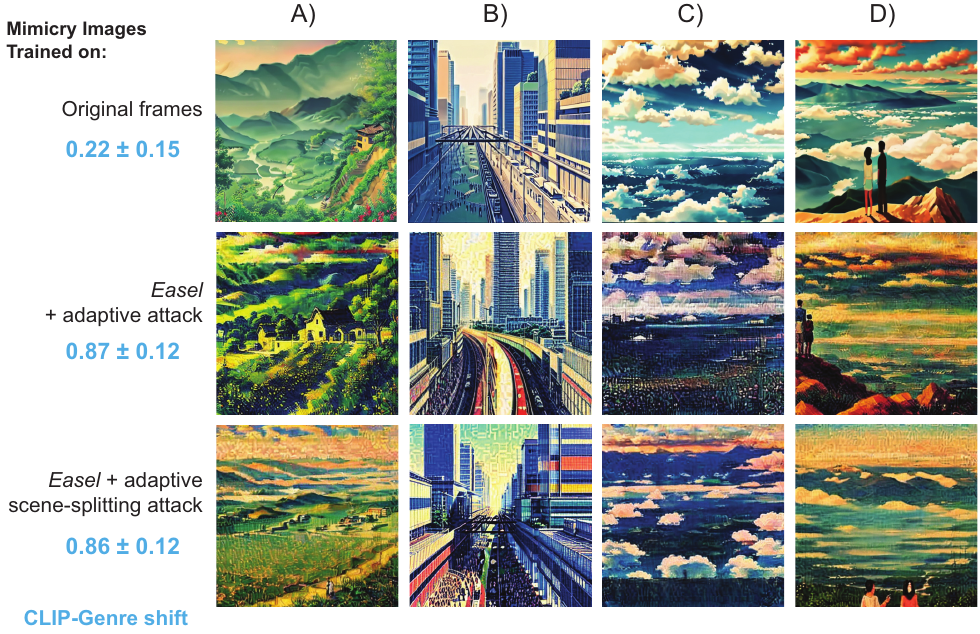}
  \caption{Visual examples of mimicry attempts on Clean frames and frames
    protected by \system~ across the standard Adaptive Attack and the
    Scene-Splitting Adaptive Attack. CLIP-Genre Shift score demonstrates
    robust protection against Mimicry attacks under both adaptive attack
    scenarios.} 
  \label{fig:style-mimicry-ctr}
\end{figure}

\para{\system~ is robust to scene-splitting attack.}
Tables~\ref{tab:countermeasure-loss-score}
and~\ref{tab:countermeasure-pd-score} show results that quantify the
image-level difference between the original frames and the frames produced by
the scene-splitting attack. They show that \system~ is robust to the
scene-splitting attack combined with pixel averaging attacks attempting to
recover original frames. There is only a minimal decrease in $L_2$ norm and
$MPD$ value when \system~ is applied to videos under the new Adaptive
Scene-Splitting Attack. In Figure~\ref{fig:style-mimicry-ctr}, we 
further verify that style mimicry using a combination of
scene-splitting and pixel-averaging is still unsuccessful, through both
visual examples of generated images and associated CLIP-genre shift scores,
which remain unchanged under the scene splitting attack.

\para{Why does the attack fail?} The failure of the scene-splitting attack
makes sense, once we consider its limitations. Regardless of where the
attacker forces a new scene break, frames inside the two new scenes ($S_1$
and $S_2$) are bounded in their maximum difference from each other. Thus the
two source frames for $S_1$ and $S_2$ (each computed as an average of frames
in the scene) is also bounded in their dissimilarity. Thus, it is likely
their resulting target tensors, and consequently the perturbations generated
from them, are also small. This intuition holds even if the attacker could
break a single long scene into several scenes of their choosing.

  \begin{table}[t]
    \centering
      \resizebox{0.5\textwidth}{!}{
      \centering
  \begin{tabular}{c|ccc}
    & Naive         & \multicolumn{1}{c}{\begin{tabular}[c]{@{}c@{}}\system\\ + Adaptive Attack\end{tabular}} & \multicolumn{1}{c}{\begin{tabular}[c]{@{}c@{}}\system~  + Adaptive \\ Scene-Splitting Attack\end{tabular}} \\ \hline
    Video Game     & 405.38 $\pm$ 19.43 & 421.21 $\pm$ 25.52 & 394.80 $\pm$ 25.16       \\
    Japanese Anime & 396.76 $\pm$ 19.05 & 406.43 $\pm$ 24.07 & 387.06 $\pm$ 23.98
\end{tabular}
    }\caption{Latent $L_2$ norm between original and perturbed frames across Video Games and Japanese Anime datasets. Demonstrates Scene-Splitting Adaptive Attack is not able to significantly reduce protection \system~ offers.}
  \label{tab:countermeasure-loss-score}
  \end{table}

  \begin{table}[t]
    \centering
      \resizebox{0.5\textwidth}{!}{
      \centering
  \begin{tabular}{c|ccc}
    & Naive         & \multicolumn{1}{c}{\begin{tabular}[c]{@{}c@{}}\system\\ + Adaptive Attack\end{tabular}} & \multicolumn{1}{c}{\begin{tabular}[c]{@{}c@{}}\system~  + Adaptive \\ Scene-Splitting Attack\end{tabular}} \\ \hline
    Video Game     & 111.58 $\pm$ 12.38 & 107.42 $\pm$ 13.47 & 100.28 $\pm$ 11.81 \\
    Japanese Anime & 112.66 $\pm$ 11.05 & 108.57 $\pm$ 12.59 & 102.02 $\pm$ 11.43       
\end{tabular}
    }\caption{MPD between original and perturbed frames across Video Games and Japanese Anime datasets. Demonstrates Scene-Splitting Adaptive Attack is not able to significantly reduce protection \system~ offers.}
  \label{tab:countermeasure-pd-score}
  \end{table}

\section{Discussion and Limitations}
\label{sec:discussion}

Our work provides a first step to addressing the rising challenge of style
mimicry through training data extracted from video imagery. However, we note
that this work still has significant limitations.

First, we note that our work primarily targets and addresses the
pixel-averaging adaptive attack. Given time, it is entirely possible for new
distinct adaptive attacks to arise in the video mimicry space that \system{}
is not designed to mitigate. Similarly, it is also possible that effective
countermeasures will be identified which target \system{} itself beyond the
scene-splitting attack. Second, while \system{} reduces computation costs
significantly over a naive application of anti-mimicry tools to each frame,
the resulting computation is still quite significant, and likely remains out
of reach for smaller video or animation creators.  Finally, it is possible
that image mimicry attacks might be supplanted altogether by some form of
video mimicry attack in the future. Such risks and potential mitigation
methods might become more clear as we learn more about generative video
models and their training process.

\bibliographystyle{ACM-Reference-Format}
\bibliography{vg}
\balance

\appendix
\section{Appendix}
\label{sec:appendix}

\subsection{Detailed Analysis on Universal Target Generation}
\label{app:detailed-universal-target}
\begin{figure}[h]
    \centering
    \includegraphics[width=0.8\columnwidth]{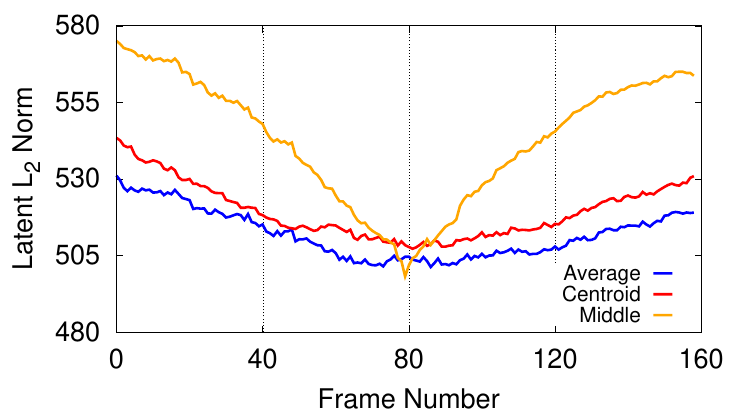}
    \vspace{-0.15in}
    \caption{Generating a single target image by pixel averaging all images in a scene leads to the least amount of fluctuation and distance in latent $L_2$ norm between each frame to the target image.}
    \label{fig:target-selection-algorithm-results}
\end{figure}

The goal of a universal target is to be sufficiently close in latent image space for each frame to optimize towards. In Figure~\ref{fig:target-selection-algorithm-results}, we plot the latent $L_2$ norm between each frame in a scene to a universal target generated by three different methods. We find that simple pixel averaging generates universal target images that are consistently closer to all frames in a scene, and thus adopt it for our evaluation section.

\begin{figure*}[t]
    \centering
    \begin{minipage}[t]{0.49\textwidth}
    \centering
    \includegraphics[width=0.9\columnwidth]{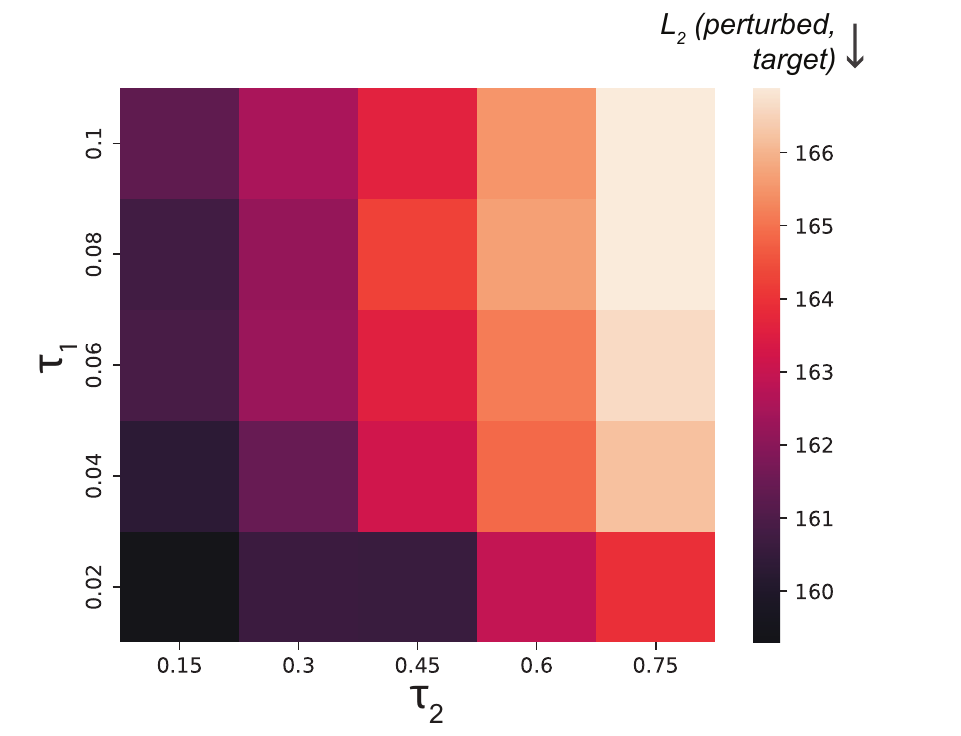}
    \vspace{-0.1in}
    \caption{Gridsearch of \system's two threshold parameters $\tau_1, \tau_2$ with respect to the latent $L_2$ norm between perturbed frames and the target it is optimized towards. A lower latent $L_2$ norm equates to better optimized perturbation and stronger robust protection. Decreasing both values leads to higher number of full optimizations, and stronger robustness.}
    \label{fig:grid-search-robustness}
    \end{minipage}
    \hfill
    \centering
    \begin{minipage}[t]{0.49\textwidth}
    \centering
    \includegraphics[width=0.9\columnwidth]{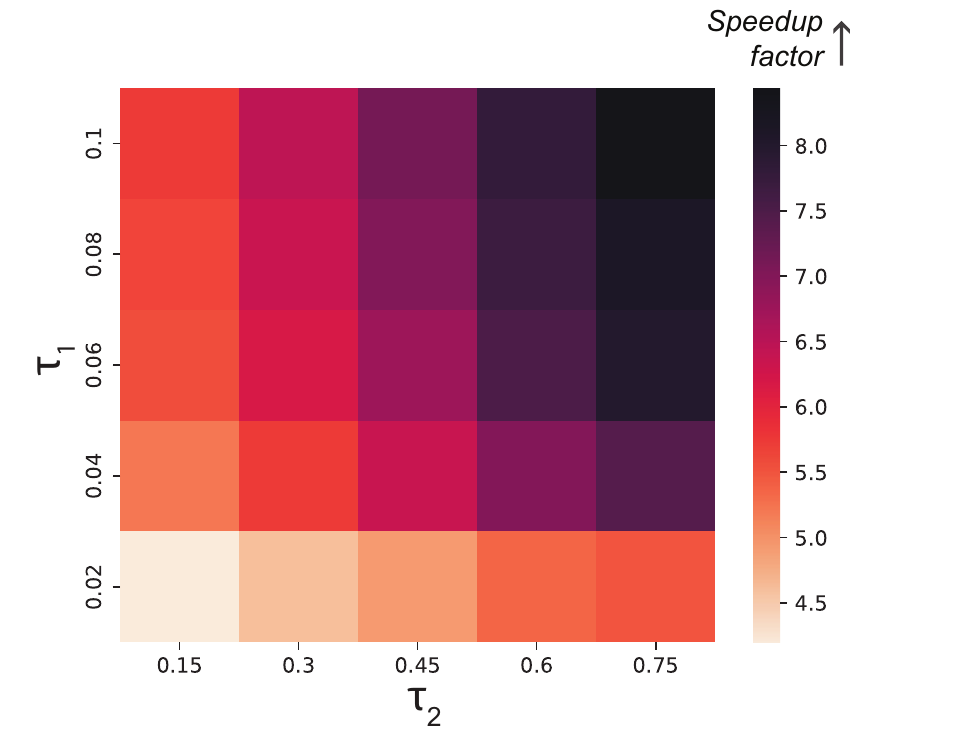}
    \vspace{-0.1in}
    \caption{Gridsearch of \system's two threshold parameters $\tau_1, \tau_2$ with respect to the computation speedup. Increasing both values leads to less number of full optimizations, and more significant speedup.}
    \label{fig:grid-search-robustness}
    \label{fig:grid-search-efficiency}
    \end{minipage}
      \hfill
\end{figure*}

\begin{figure*}[t]
  \centering
  \begin{minipage}[t]{0.32\textwidth}
  \centering
  \includegraphics[width=0.9\columnwidth]{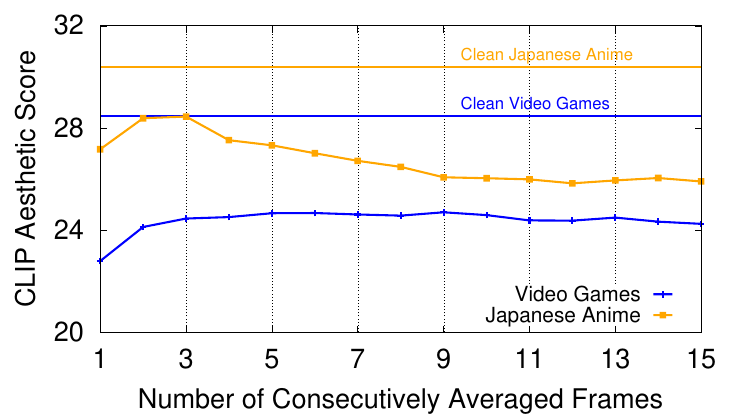}
  \vspace{-0.1in}
  \caption{Average image quality (CLIP Aesthetic score) increases initially as the number of consecutively frames used for averaging attack increases, but decreases after too many frames are averaged for both Video Game and Japanese Anime videos.}
  \label{fig:aesthetic-caf}
  \end{minipage}
  \hfill
  \centering
  \begin{minipage}[t]{0.32\textwidth}
  \centering
  \includegraphics[width=0.9\columnwidth]{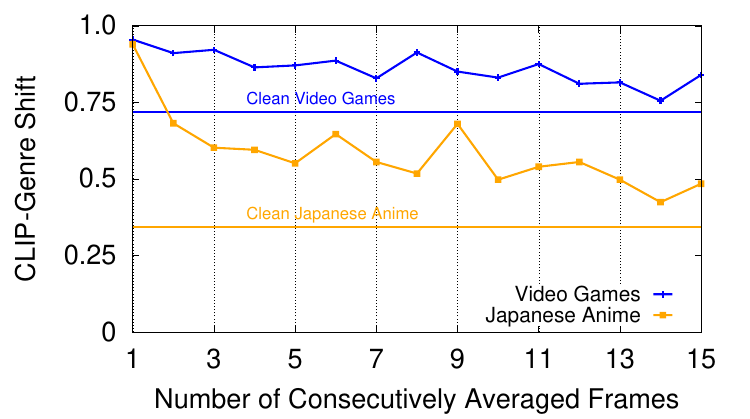}
  \vspace{-0.1in}
  \caption{Average robustness (CLIP-Genre Shift) decreases as the number of consecutively frames used for averaging attack increases and stagnates as more frames are used for both Video Game and Japanese Anime videos.}
  \label{fig:clip-shift-caf}
  \end{minipage}
    \hfill
    \centering
  \begin{minipage}[t]{0.32\textwidth}
  \centering
  \includegraphics[width=0.9\columnwidth]{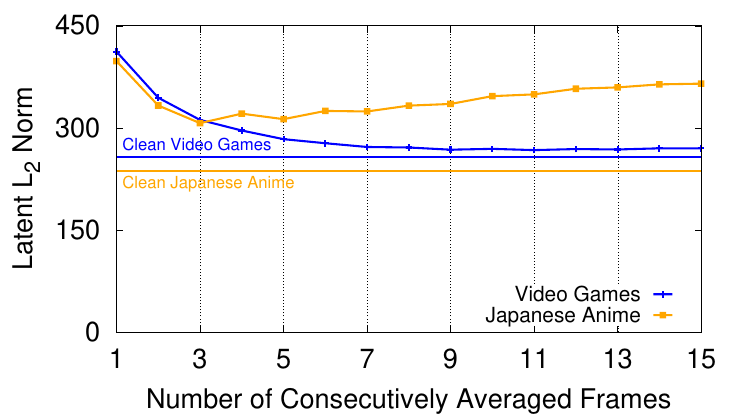}
  \vspace{-0.1in}
  \caption{Average latent $L_2$ norm between perturbed frames and clean frames initially decreases as the number of consecutive frames used for averaging attack increases. The benefit stagnates for Video Game videos, but loses effectiveness for Japanese Anime.}
  \label{fig:l2norm-caf}
  \end{minipage}
    \hfill
\end{figure*}

\subsection{Detailed Perturbation Removal Attacks}
\label{app:detailed-perturbation-removal}

\para{FILM}
Frame Interpolation for Large Motion (FILM)~\cite{reda2022film} is a neural network designed to generate slow motion videos from two ``near duplicate'' images. FILM adapts a multi-scale feature extractor to estimate scale-agnostic motion, allowing the synthesis of high quality intermediate frames. While there are many SOTA film interpolation softwares available ~\cite{niklaus2020softmax, bao2019depth, huang2022real}, these focus largely on estimating motion and optical flow. In contrast, FILM prioritizes generating sharp, high quality videos by using Gram matrix loss ~\cite{gatys2016image} to fill in the gap between two or more images. We implement their code to generate a single interpolated image between two images separated five frames apart. FILM is designed to handle both large and small motion, thus we find that the distance between frames does not impact the success of this removal attack.

\para{Simple Linear Interpolation}
We apply a simple linear interpolation function to generate a new image by smoothly blending the pixel values of two input images. We chose an interpolation factor $\alpha = 0.5$ to give equal weight to both images. Similar to our existing perturbation removal attacks, we perform simple linear interpolation between two images 5 frames apart. 

\subsection{Number of Images to use in Average Attack}
\label{app:num-images-average}
Figures \ref{fig:aesthetic-caf}-\ref{fig:l2norm-caf} plot the robustness and quality of protected images as a function of the number of consecutive frames used by our selective pixel averaging attack. Image quality begins to degrade when more than 3-5 consecutive frames are used for selective pixel averaging, with diminishing returns on recovering the correct artist style. Thus, we fix the number of averaged frames to 5 throughout our evaluation.

\subsection{User Study Detailed Description}
\label{app:detailed-user-study}
\para{Ethics}
We conduct two IRB-approved user studies. The first study, released on Twitter, involved 305 volunteers from the art community who rated the effectiveness of \system~ compared to naive perturbations in protecting videos against style mimicry attempts. We only presented style-mimicked images from the Video Game and Japanese Anime datasets, chosen due to their relevance to current threats faced by video creators as discussed in \S\ref{sec:intro}. These datasets align with art content that existing image protection systems (Glaze, Mist, AntiDB \cite{shan2023glaze,mist,antidb}) are designed to safeguard.

The second study, conducted on Prolific, involved 220 participants who were presented with style-mimicked images from videos across all 5 datasets. They were compensated at \$12 per hour and filtered for English-Speaking participants in the US, over the age of 18 and with over 95\% approval rating on Prolific.

All videos from these datasets are publicly available on YouTube, and we strictly use them for research purposes.

\subsection{Choosing System Parameters}
\label{app:system-parameters}
This is a detailed explanation on the selection of two threshold parameters introduced in \S\ref{subsec:system-parameters}. We perform a grid search on 25 combinations of \system's two threshold parameters using Glaze as the image perturbation algorithm, and measure the average robustness and speedup that each achieves. In Figures \ref{fig:grid-search-robustness} and \ref{fig:grid-search-efficiency}, we find that there is an intuitive reverse tradeoff between robustness and computation efficiency. Decreasing both parameters leads to more computations of perturbations from scratch, which benefits robustness but incurs slower speedup. Assuming that video content creators must balance these two objectives, we choose the middle parameters, $\tau_1 = 0.06$ and $\tau_2 = 0.45$, to use in our evaluation.

\subsection{Averaging Images Degrades Mimicry Quality}
Here, we investigate the impact that our averaging attack on style mimicry models trained on clean images. We take clean unperturbed frames of four videos from the Japanese Anime dataset and perform our averaging attack on the 5 frames surrounding each high quality frame used in style mimicry. Our results show that the average attack on clean images causes CLIP-Genre Shift to increase from 0.15 to 0.19. This suggest that our average attacks slightly weakens style mimicry, which is supported by natural blurring caused by movement and inter-frame changes. Additionally, this result helps explain why averaging attack on \system~ has higher CLIP-Genre Shift scores when compared to naive perturbation methods. \system~ successfully protects against averaging attack, but style mimicry is also compounded by the natural degradation of image quality when averaging attack is used.

\subsection{Increasing $\tau_2$ Degrades Robustness}
\label{app:eps-robustness}
In \S\ref{subsec:protection-usability}, we show that speedup from \system~ can be significantly improved by increasing our second threshold parameter $\tau_2$. While our grid search signals that increasing either threshold parameter leads to lower robustness via. latent $L_2$ norm, we also show that increasing $\tau_2$ leads to worse robustness on end to end style mimicry models. On the same four Japanese Anime videos used to evaluate computation efficiency in \S\ref{subsec:protection-usability}, the default $\tau_2 = 0.45$ results in average CLIP-Genre Shift score of 0.95, while increasing $\tau_2 = 0.8$ drops CLIP-Genre Shift score to 0.90.

\end{document}